\documentclass[11pt,letterpaper]{article}
\usepackage{sson_arxiv}

\usepackage{graphicx}
\usepackage{amsmath}
\usepackage{booktabs}
\usepackage{array}
\usepackage{placeins} 
\usepackage{float} 
\usepackage{enumitem}
\usepackage{hyperref}
\usepackage{url}
\usepackage{fvextra} 
\usepackage{tabularx,xcolor,pifont,caption}
\usepackage{trimclip}
\newsavebox{\ssonFrameworkBox}

\title{Same Space or Not: Benchmarking Spatial\\
Intelligence Through Cross-View Contradictions}

\newcommand{\SSoNAuthors}{%
  Huizi Cao\textsuperscript{1}, Guanyu Xu\textsuperscript{2},
  Zhixuan Liang\textsuperscript{3}, Ming Yin\textsuperscript{3},
  Jiuxiang Gu\textsuperscript{4}, Manling Li\textsuperscript{5},
  Mengdi Wang\textsuperscript{3}, Shilong Liu\textsuperscript{3}}
\newcommand{\SSoNAffiliations}{%
  \textsuperscript{1}Carleton College\quad
  \textsuperscript{2}University of Pennsylvania\quad
  \textsuperscript{3}Princeton University\quad
  \textsuperscript{4}Adobe Research\quad
  \textsuperscript{5}Northwestern University}
\newcommand{\SSoNAuthorNote}{}

\newcommand{\SSoNDatasetURL}{https://huggingface.co/datasets/HuiziCao/SSoN}
\newcommand{\SSoNProjectURL}{http://sson.aisexam.com/}
\newcommand{\SSoNCheckpointsURL}{https://huggingface.co/HuiziCao/SSoN-Qwen2.5-VL-7B-LoRA}
\newcommand{\SSoNGitHubURL}{https://github.com/sson-benchmark/SSoN}

\author{\SSoNAuthors}
\hypersetup{pdfauthor={Huizi Cao, Guanyu Xu, Zhixuan Liang, Ming Yin, Jiuxiang Gu, Manling Li, Mengdi Wang, Shilong Liu}}

\newcommand{\bench}{SSoN}

\newcommand{\myparagraph}[1]{\textbf{#1}}

\begin{document}

\maketitle

\begin{abstract}
Spatial intelligence, the capacity to represent and reason about physical space, underlies embodied perception and real-world visual understanding. Existing benchmarks probe it through spatial relations, perspective-taking, and mental modeling.
What remains underexamined is whether a model can maintain a coherent scene representation across views, since pairwise comparison alone cannot reveal which view departs from the truth.
We thus introduce Same Space or Not (SSoN), a benchmark of cross-view contradiction resolution comprising 1,000 questions from seven indoor scene sources and six categories of controlled 3D edits. 
Each question presents three views of an unchanged scene and one view of an edited variant. Models must identify the inconsistent view and explain the contradiction without a designated trusted reference or a marked target. 
Our construction process iteratively translates human review into generation constraints. Automated checks and exhaustive human validation screen out both trivial visual shortcuts and questions lacking sufficient evidence for a unique answer. 
Humans achieve 83.3\% accuracy, compared with 39.7\% for the strongest vision-language models (VLMs), which further drops to 34.9\% when requiring valid visual explanation for the correct selection.
We investigate prompting, camera information, and 3D reconstruction as routes for improvement.
On a 100-question intervention set with a 24\% baseline, prompting and 3D reconstruction remain below 50\% answer accuracy, except when ground-truth camera information is supplied, which reaches 61\% answer accuracy but only 48\% True solve accuracy.
On a held-out 201-question test set, Qwen2.5-VL-7B achieves 21.89\% accuracy, 33.33\% after answer-only fine-tuning, and 41.79\% after fine-tuning with both rationale and answer supervision.
These results identify cross-view contradiction resolution as a substantial weakness of current VLMs that can be partly addressed by targeted supervision. The 1,000 questions and fine-tuned checkpoints will be made public.
\end{abstract}

\section{Introduction}
\label{sec:intro}

\begingroup
\setlength{\intextsep}{3pt}
\begin{figure}[H]
    \centering
    \includegraphics[width=0.977\linewidth]{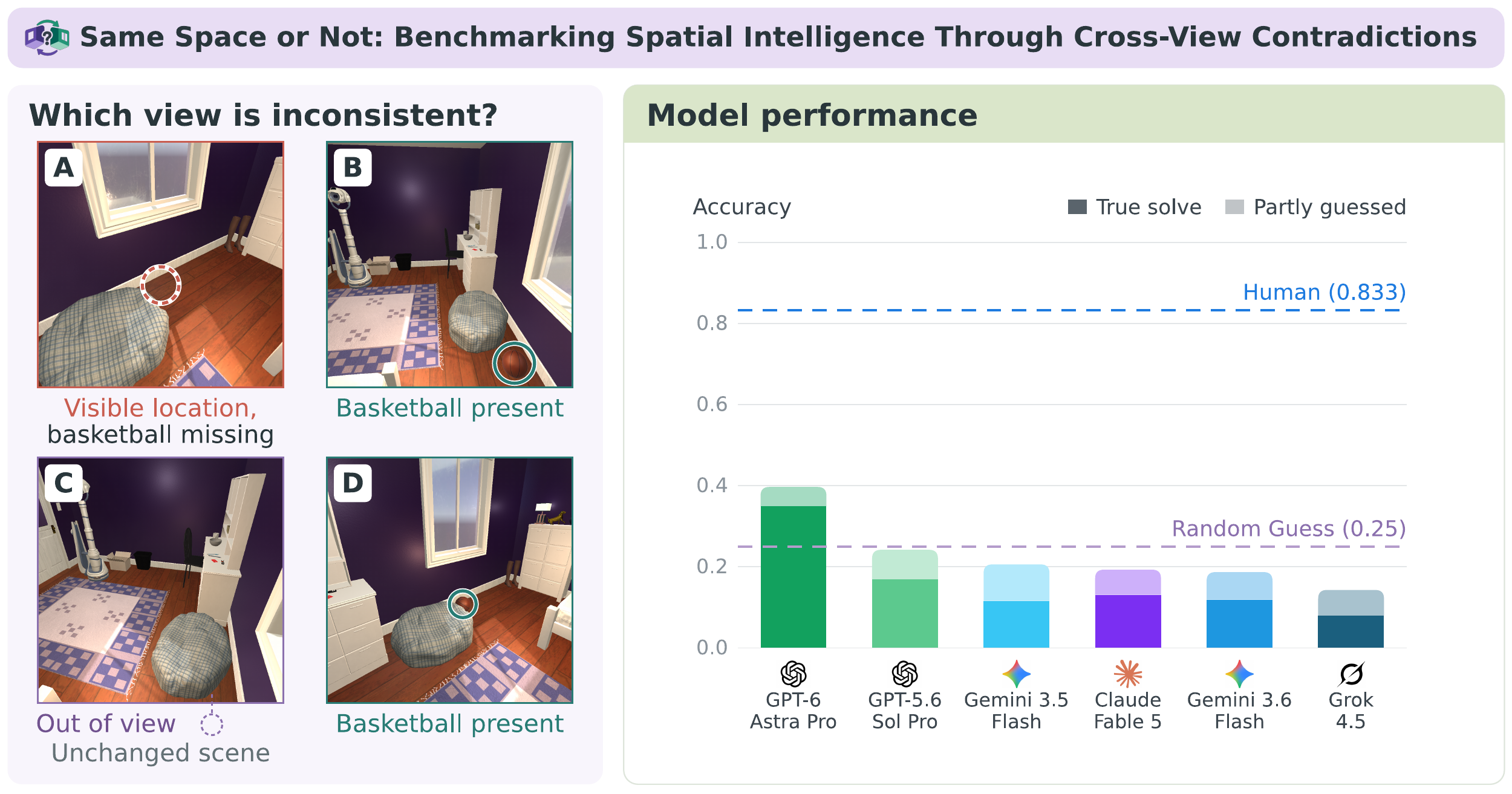}
    \caption{\textbf{SSoN task example and model performance.}
    Models identify the view that contradicts the scene depicted
    by the other three. Stacked bars separate correct selections
    into True solve and Partly guessed. Dashed lines indicate
    human accuracy and the random guessing level.}
    \label{fig:teaser}
\end{figure}
\endgroup
\clearpage

Spatial intelligence is essential for understanding and acting in the physical world. For vision-language models (VLMs), a key dimension of capabilities is understanding of \emph{spatial consistency}, which means maintaining a coherent account of the world across viewpoints rather than interpreting images in isolation~\citep{wang2026mindcube,yang2026mmsi}. Camera motion can change appearance, projected position, and even visibility of certain perspectives without changing the scene, while an edited scene may remain plausible in every individual image. Therefore, a VLM model must be able to distinguish changes in observation viewpoint from changes in the scene. This is a judgment that requires reconciling evidence across views, not merely detecting visual differences.

Existing benchmarks assess spatial relations, perspective-taking, and mental modeling across views~\citep{yang2026mmsi,zhang2026spinbench,wang2026mindcube}, and recent work extends this line to detecting inconsistency within a single pair of views~\citep{khangaonkar2026inconsistencies}. However, a pairwise comparison can only flag that two views disagree. It cannot determine which one is unfaithful to the underlying scene, since either view could equally be the anomalous one when absent further evidence. Detecting a conflict is therefore not the same as resolving it. We introduce \emph{Same Space or Not} (\bench{}) to study \emph{set-level consistency attribution}.
Given a set of views in which exactly one is inconsistent with the rest, a model must identify the culprit view and justify its choice using corroborating evidence from the remaining views, without access to a designated trusted reference or a pre-identified changed object.

We construct 1,000 four-view items from seven virtual indoor scene sources. Each contains three views of an unchanged scene and one view rendered after a controlled edit drawn from six categories (\emph{swap}, \emph{move}, \emph{rotate}, \emph{mirror}, \emph{delete}, and \emph{detail}). Edit magnitudes, object sizes, and camera viewpoints vary the reasoning demands. We additionally introduce distractor views without scene changes to incorporate apparent visual differences through occlusion and newly visible content, challenging models to distinguish viewpoint effects from real scene changes.

Besides, construction must preserve the need for genuine spatial reasoning while providing sufficient evidence for a unique answer. We enforce these requirements through \emph{validity-guided constraint discovery}, an iterative process that converts failure cases identified during human review into reusable generation and verification rules, together with automated checks and exhaustive human review applied to every item. Because construction guarantees that such evidence exists, evaluation likewise checks the evidence behind a correct selection. We distinguish \emph{True solve} from \emph{Partly guessed} outcomes according to whether a model's explanation provides factually valid, sufficient support for the answer, rather than crediting a correct choice reached without such support.

For this benchmark, humans achieve 83.3\% accuracy, compared with 39.7\% for the strongest of six VLMs (Figure~\ref{fig:teaser}). Similar overall scores conceal uneven gains across model variants, and hallucinated explanations reduce the strongest model's True solve accuracy to 34.9\%. On a separate 100-item subset with a 24\% baseline, prompting and 3D reconstruction remain below 50\% answer accuracy. The exception is the condition with explicitly supplied ground-truth camera information, which achieves 61\% answer accuracy but only 48\% True solve accuracy. And additional training provides another route to improve. Using a 700/99/201 train/validation/test split with rooms kept disjoint across partitions, Qwen2.5-VL-7B improves from 21.89\% to 33.33\% with answer supervision and 41.79\% with additional rationale supervision.
Thus, our contributions are threefold.
\begin{itemize}[itemsep=0pt, topsep=0pt]
    \item \textbf{A benchmark for spatial intelligence.}
    We present \bench{}, which targets spatial intelligence through contradiction attribution across controlled
    3D scene edits, spanning 1,000 items drawn from seven scene sources.

    \item \textbf{Validity-guided constraint discovery.}
    Building \bench{} requires reconciling two competing goals, preserving the need for spatial reasoning while ensuring sufficient evidence for a unique answer. We distill this process into reusable construction rules.

    \item \textbf{Diagnosis and improvement studies.}
    Using \bench{}, we quantify the gap between humans and models, expose hallucinated explanations, and evaluate prompting, reconstruction, and task-specific supervision.

\end{itemize}



\renewcommand{\myparagraph}[1]{\textbf{#1}}
\captionsetup[table]{skip=3pt}
\section{Related Work}
\label{sec:related}

\myparagraph{Spatial intelligence across observations.}
3DSRBench evaluates 3D relations and viewpoint robustness~\citep{ma2025threeD}, while SpinBench probes rotation and perspective-taking~\citep{zhang2026spinbench}. VSI-Bench tests spatial recall from video~\citep{yang2025vsi}, and MMSI-Bench~\citep{yang2026mmsi} and MindCube~\citep{wang2026mindcube} evaluate spatial relations and mental modeling across views. DSI-Bench examines how models distinguish observer motion from object motion~\citep{zhang2025dsi}, while ESI-Bench studies the acquisition of informative observations through movement and interaction~\citep{hong2026esi}. \bench{} makes the compatibility of observations itself the prediction target. Given sufficient visual evidence, a model must identify the view that contradicts the scene supported by the others. This isolates the ability to reconcile scene states across viewpoints from action selection and continuous temporal tracking.

\myparagraph{From scene changes to contradiction attribution.}
MOCHI~\citep{bonnen2024mochi} evaluates multiview object consistency through triplets containing two views of one object and one view of another, with the task of identifying the different object. CLEVR-Change separates scene edits from camera and illumination changes through paired-image captioning~\citep{park2019change}. MindEdit-Bench queries the consequences of hypothetical object edits without showing their resulting configurations~\citep{yu2026mindedit}. Paired inconsistency detection asks models to select a labeled object altered through cross-view image compositing~\citep{khangaonkar2026inconsistencies}. XVR-Eval identifies views with inconsistent correspondence markers or timestamps~\citep{jeong2026xvr}. \bench{} turns controlled interventions in the underlying 3D scene into a task of identifying and explaining an incompatible view. The model must discover an unmarked discrepancy, resolve it through corroborating observations, and support its selection with valid visual evidence. Table~\ref{tab:benchmark_comparison} highlights this combination of scene intervention, view attribution, and joint evaluation.

\myparagraph{Validity and explanatory evidence.}
MMSI-Bench uses annotated rationales to diagnose spatial errors~\citep{yang2026mmsi}, and ESI-Bench verifies answerability and non-triviality~\citep{hong2026esi}. MindEdit-Bench likewise applies human verification to automatically generated spatial questions~\citep{yu2026mindedit}. \bench{} connects construction validity and response evaluation through the same underlying scene intervention. Our validity-guided constraint discovery translates review failures into generation and verification rules that preserve the need for spatial reasoning while retaining sufficient evidence for a unique answer. Both automated checks and exhaustive human review enforce these requirements. Joint scoring then separates \emph{True solve} from \emph{Partly guessed} outcomes among correct selections, exposing hallucinated or insufficient explanations that choice accuracy alone cannot detect.

\begin{table}[H]
\centering
\caption{\textbf{Comparison with previous benchmarks.} Rendered edits change a shared 3D scene. Joint scoring requires valid evidence for a correct choice. The last column requires reported item-level human review of answer uniqueness and visual shortcuts. Crosses in that column mean this joint audit is unreported.}

\label{tab:benchmark_comparison}
\begingroup
\small
\definecolor{ssonCheckColor}{HTML}{2E8B57}
\definecolor{ssonCrossColor}{HTML}{999999}
\newcommand{\ssonYes}{\textcolor{ssonCheckColor}{\ding{51}}}
\newcommand{\ssonNo}{\textcolor{ssonCrossColor}{\ding{55}}}
\setlength{\tabcolsep}{3pt}
\renewcommand{\arraystretch}{1.05}

\begin{tabularx}{\linewidth}{@{}
    >{\raggedright\arraybackslash}p{0.34\linewidth}
    *{5}{>{\centering\arraybackslash}X}@{}}
\toprule
\textbf{Benchmark}
& \shortstack{Multiple\\viewpoints}
& \shortstack{Rendered\\scene edits}
& \shortstack{Intruder\\view\\selection}
& \shortstack{Joint\\answer and\\evidence\\scoring}
& \shortstack{Validity \&\\shortcut\\audit} \\
\midrule
VSI-Bench~\citep{yang2025vsi}
& \ssonYes & \ssonNo & \ssonNo & \ssonNo & \ssonNo \\
MindEdit-Bench~\citep{yu2026mindedit}
& \ssonYes & \ssonNo & \ssonNo & \ssonNo & \ssonNo \\
CLEVR-Change~\citep{park2019change}
& \ssonYes & \ssonYes & \ssonNo & \ssonNo & \ssonNo \\
Paired inconsistency\newline
\citep{khangaonkar2026inconsistencies}
& \ssonYes & \ssonNo & \ssonNo & \ssonNo & \ssonNo \\
MOCHI~\citep{bonnen2024mochi}
& \ssonYes & \ssonNo & \ssonYes$^{\dagger}$ & \ssonNo & \ssonNo \\
XVR-Eval~\citep{jeong2026xvr}
& \ssonYes & \ssonNo & \ssonYes & \ssonNo & \ssonNo \\
\midrule
\textbf{\bench{} (ours)}
& \ssonYes & \ssonYes & \ssonYes & \ssonYes & \ssonYes \\
\bottomrule
\end{tabularx}
\endgroup
\vspace{3pt}

\footnotesize\raggedright
$\dagger$ For MOCHI, intruder selection identifies the different object in a multiview triplet. The MindEdit-Bench and XVR-Eval rows cover counterfactual editing and verification tasks, respectively. 
\vspace{-2pt}
\end{table}

\section{Methods and Benchmark}
\label{sec:bench}
\definecolor{ssonTrueBG}{HTML}{E6EEDC}
\newcommand{\ssonTS}[1]{\begingroup\setlength{\fboxsep}{0pt}\colorbox{ssonTrueBG}{\makebox[2.6em][r]{\strut#1\hspace{0.8pt}}}\endgroup}

\begin{figure}[H]
    \centering
    \includegraphics[width=1.0\textwidth]{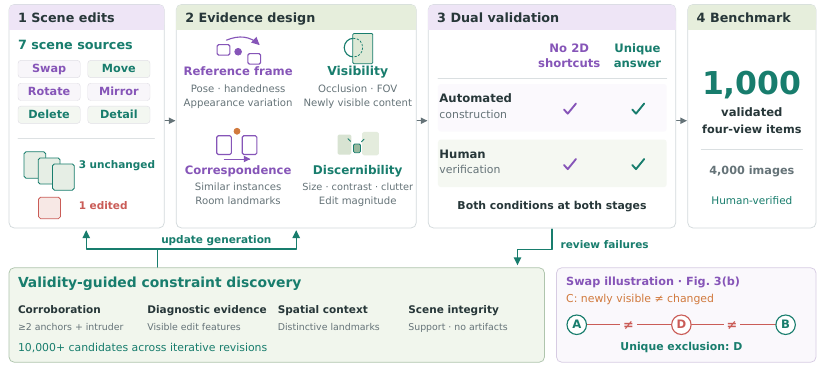}
    \vspace{-4pt}
    \caption{\textbf{Validity-guided benchmark construction for \bench{}.}}
    \label{fig:framework}
    \vspace{-6pt}
\end{figure}

We jointly design scene edits and viewpoints to meet two competing requirements, preserving spatial reasoning and ensuring a unique answer. Direct appearance comparisons can permit shortcuts, whereas excessive occlusion or viewpoint separation can remove decisive evidence. We therefore construct corroborating observations alongside valid distractors that increase reasoning demands while preserving answerability (Figure~\ref{fig:framework}).

\begin{figure}[p]
    \centering
    \includegraphics[width=1.0\linewidth]{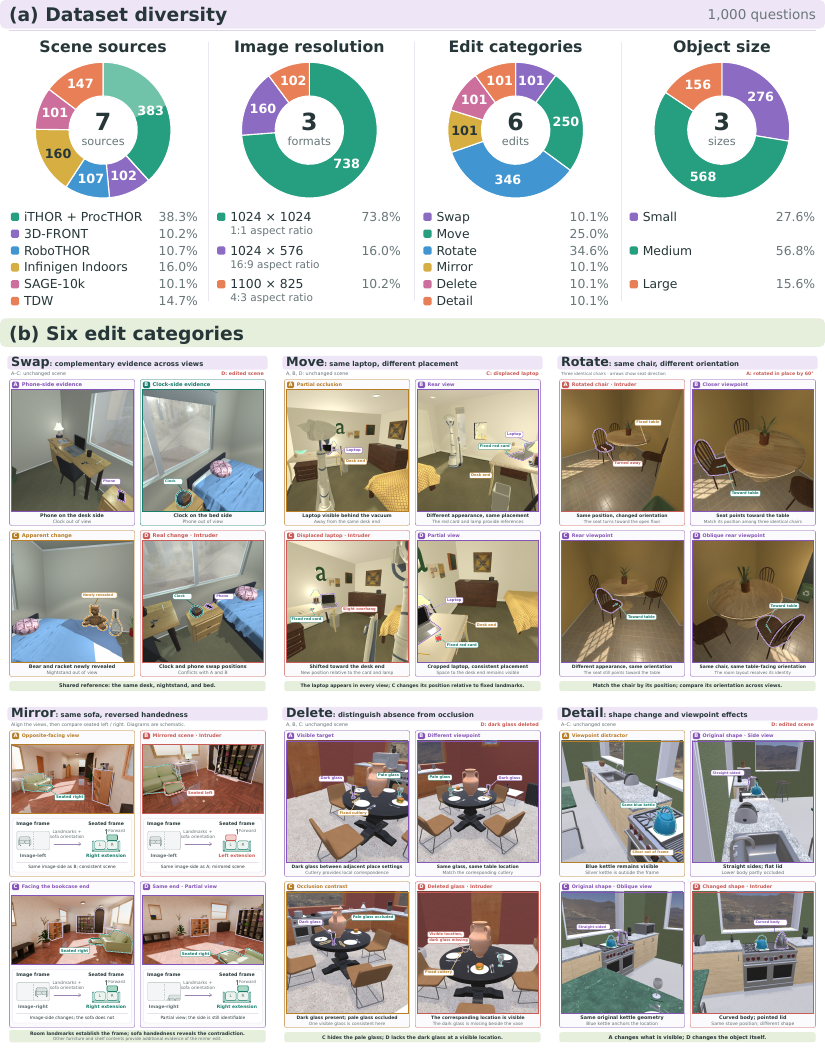}
    \caption{\textbf{Dataset diversity and representative questions in \bench{}.}
    (a) Sources, resolutions, edit categories, and object sizes.
    (b) Examples across six edits. Evidence annotations are excluded from standard evaluation.}
    \label{fig:framework_detail}
    \vspace{-6pt}
\end{figure}

\subsection{Validity-guided Benchmark Construction}
\label{sec:pipeline}

\myparagraph{Defining the scene change.}
An edit of category $\tau$ transforms scene $S$ into $S'=T_{\tau}(S)$. With intruder index $y\in\{A,B,C,D\}$, we render
\begin{equation}
 I_i=\begin{cases}
 \mathcal{R}(S',c_i), & i=y,\\
 \mathcal{R}(S,c_i), & i\ne y,
 \end{cases}
 \qquad i\in\{A,B,C,D\},
 \label{eq:item_construction}
\end{equation}
where $c_i$ specifies the camera and $\mathcal{R}$ the source-specific renderer. The three unchanged views are \emph{anchors}. Models identify and explain the intruder from these images and the prompt alone, without a trusted reference or marked target. Cameras must make the edit attributable.

\emph{Move} includes chair \emph{pullout} and \emph{quarter}, \emph{half}, and \emph{full} translations with branch-dependent distances. \emph{Rotate} applies $30^\circ$, $45^\circ$, $60^\circ$, or $90^\circ$ to unique objects or repeated instances, with auxiliary translation for feasible placement in some THOR items. \emph{Swap} exchanges positions and optionally orientations, aligning incompatible facing directions to prevent conspicuous front-to-back shortcuts. \emph{Mirror} reflects the room configuration while retaining its objects. \emph{Delete} removes an object. \emph{Detail} changes identity, shape, or appearance at a corresponding location through replacement, recoloring, or retexturing, including cross-category replacements.

\myparagraph{Retaining sufficient visual evidence.}
The intruder and at least two anchors must expose usable diagnostic features in the edited region at a discernible scale with environmental references. Automated image-area checks screen candidates. Human reviewers judge discernibility and frame-edge blending. Informative partial views can qualify. The third anchor may corroborate or distract. For delete, pre-edit visibility from the intruder camera and evidence at the expected location must exclude occlusion or framing as explanations for absence.

\myparagraph{Challenging shortcuts through apparent changes.}
Unchanged same-room distractors resemble edits, making object presence or appearance unreliable for selection. They challenge shortcuts and increase cross-view reasoning demands while retaining room cues and potentially diagnostic evidence, without relaxing the two-anchor requirement.

\myparagraph{Validity-guided constraint discovery.}
Review of over 10,000 development candidates converted insufficient corroboration, barely visible edits, disorienting views, and physical defects into witness, image-scale, room-reference, and placement requirements, respectively. Model feedback and development studies (Section~\ref{sec:difficulty}) informed revisions. Generation, automated screening, and human review apply common criteria through source- and version-specific implementations. Final assembly rejected invalid items and balanced coverage before separately evaluating all 1,000 questions.

\subsection{Human Verification}
\label{sec:human_verification}

Candidates pass three human checks before entering the bank. First, reviewers use A--D, the key, and paired top-down views to verify physical integrity and diagnostic evidence in the intruder and two anchors. Second, a reviewer new to the item records a choice, evidence, and ambiguity from A--D before seeing the key, then checks for alternative answers. Confirmed errors or ambiguity cause rejection. Third, reviewers test isolated appearance, inventories, and nearly aligned pixel comparisons as shortcuts. The final 1,000 items were frozen before human and model evaluation, with no later deletions or replacements. These checks are separate from performance testing and explanation grading (Appendix~\ref{app:construction_audit}).

\subsection{Benchmark Statistics}
\label{sec:stats}

Seven indoor sources contribute 1,000 questions (Figure~\ref{fig:framework_detail}(a)). Blender renders 3D-FRONT~\citep{fu2021front}, Infinigen Indoors~\citep{raistrick2024indoors}, and SAGE-10k~\citep{xia2026sage}. Unity-based environments provide iTHOR~\citep{kolve2017thor}, ProcTHOR~\citep{deitke2022procthor}, RoboTHOR~\citep{deitke2020robothor}, and TDW~\citep{gan2021tdw}. Shared room configurations and components prevent consistent separation of iTHOR and ProcTHOR, reported jointly as 383 questions, with 107 separate RoboTHOR questions.

Each edit category has over 100 questions. Move and rotate account for 25.0\% and 34.6\%. Size uses the unchanged object's longest axis-aligned bounding-box side $d$ after scene scaling. Small ($d<0.3\,\mathrm{m}$), medium ($0.3\le d<1\,\mathrm{m}$), and large ($d\ge1\,\mathrm{m}$) groups contain 276, 568, and 156 items. Swap uses the larger object. Mirror uses the largest object providing a principal diagnostic cue, making all retained mirror items large.

Original aspect ratios are preserved at $1024\times1024$ for 738 items, $1024\times576$ for 160 Infinigen Indoors items, and $1100\times825$ for 102 3D-FRONT items. Viewpoint separation, occlusion, instance multiplicity, and distracting content vary reasoning demands under common verification criteria without forming a fully crossed design.

\subsection{Difficulty Analysis}
\label{sec:difficulty}

\myparagraph{Relating evidence across viewpoints.}
Low contrast complicates localization and boundary separation, while repeated objects require landmark-based correspondence. Unique targets ease matching but can still require geometric reasoning. Camera motion changes projected position and apparent orientation, so move and rotate require scene-relative comparison. Swap binds identities to locations, while mirror uses asymmetry to distinguish reflection from viewpoint change. Witnesses may be redundant or complementary, while viewpoint effects themselves distract in rotate and mirror.

Distractors have three overlapping effects. Framing or occlusion causes \emph{non-observation}, additional objects amid scene overlap produce \emph{newly revealed content}, and unchanged objects exhibit \emph{viewpoint/appearance variation}. In Figure~\ref{fig:teaser}, A and C both lack the basketball, but only A exposes its expected location. B and D confirm its presence, so A is inconsistent and C valid.

\myparagraph{Development evidence for distractor design.}
A 100-item development study tested five VLMs across sources and edits before large-scale construction. Astra Pro was evaluated later on the same archived 100 items. Without designed distractors, errors were roughly balanced across anchors. With distractors, their share of errors exceeded either ordinary anchor's by over 30 percentage points. This measures error concentration rather than accuracy loss.

For detail, delete, swap, and move, a clear frontal anchor replaced the distractor, preserving the edit, intruder, other two anchors, and task while removing distraction and adding a direct reference. Astra Pro reached 100\% on delete and swap and near-perfect detail accuracy. Sol Pro, Gemini 3.5, Fable 5, and Gemini 3.6 each reached at least 80\% in all three categories, while Grok at least doubled each score. On move, Astra rose from below 30\% to above 60\%, and every other model at least doubled. These development answer accuracies are distinct from True solve, final scores, and the later 100-item intervention study.

With edits fixed, these gains distinguish recognizing an exposed change from attributing contradictions across viewpoints. Move remains harder for Astra even with a direct reference. Findings from the initial five-model study guided distractor design across edits to increase evidence-integration demands. Detection, matching, and reconstruction remain legitimate strategies when supported by visual evidence.

\FloatBarrier
\section{Experiments}
\label{sec:experiments}

We assess cross-view contradiction identification and explanation and test appearance comparison, inference guidance, geometric support, and training. Evaluation covers all 1,000 questions. Inference interventions share a 100-question subset, while fine-tuning uses a separate 201-question test split. These studies test spatial consistency's difficulty and responsiveness to intervention.

\begin{table}[H]
\centering
\caption{\textbf{Answer accuracy (Acc) and True solve (TS, green) on all 1,000 questions (\%).} Both metrics use all category questions as denominators. Bold marks the best VLM per category and metric. Human scores report Acc only.}
\label{tab:main_choice}
\label{tab:main_true}
\begingroup
\small
\setlength{\tabcolsep}{0.6pt}
\renewcommand{\arraystretch}{1.05}
\begin{tabular*}{\linewidth}{@{\extracolsep{\fill}}l|*{6}{rr|}r@{}}
\toprule
Category ($n$) & \multicolumn{2}{c|}{\shortstack{GPT-6\\Astra Pro}} & \multicolumn{2}{c|}{\shortstack{GPT-5.6\\Sol Pro}} & \multicolumn{2}{c|}{\shortstack{Gemini 3.5\\Flash}} & \multicolumn{2}{c|}{\shortstack{Claude\\Fable 5}} & \multicolumn{2}{c|}{\shortstack{Gemini 3.6\\Flash}} & \multicolumn{2}{c|}{\raisebox{5pt}[0pt][0pt]{Grok 4.5}}
& \raisebox{5pt}[0pt][0pt]{Human} \\
 & Acc & \ssonTS{TS} & Acc & \ssonTS{TS} & Acc & \ssonTS{TS} & Acc & \ssonTS{TS} & Acc & \ssonTS{TS} & Acc & \ssonTS{TS} & Acc \\
\midrule
Overall (1,000) & \textbf{39.70} & \ssonTS{\textbf{34.90}} & 24.20 & \ssonTS{16.90} & 20.60 & \ssonTS{11.60} & 19.30 & \ssonTS{13.10} & 18.70 & \ssonTS{11.90} & 14.30 & \ssonTS{8.00} & 83.30 \\
Swap (101) & \textbf{92.08} & \ssonTS{\textbf{85.15}} & 58.42 & \ssonTS{55.45} & 46.53 & \ssonTS{37.62} & 39.60 & \ssonTS{31.68} & 34.65 & \ssonTS{33.66} & 22.77 & \ssonTS{19.80} & 97.03 \\
Move (250) & \textbf{27.20} & \ssonTS{\textbf{20.40}} & 14.40 & \ssonTS{4.00} & 13.20 & \ssonTS{2.00} & 14.00 & \ssonTS{7.20} & 16.00 & \ssonTS{3.20} & 12.00 & \ssonTS{3.20} & 71.60 \\
Rotate (346) & 7.23 & \ssonTS{\textbf{5.20}} & 6.36 & \ssonTS{1.16} & \textbf{9.54} & \ssonTS{1.16} & 5.20 & \ssonTS{0.29} & 5.78 & \ssonTS{1.45} & 7.80 & \ssonTS{3.18} & 76.01 \\
Mirror (101) & \textbf{64.36} & \ssonTS{\textbf{56.44}} & 24.75 & \ssonTS{15.84} & 19.80 & \ssonTS{11.88} & 24.75 & \ssonTS{18.81} & 23.76 & \ssonTS{19.80} & 23.76 & \ssonTS{13.86} & 95.05 \\
Delete (101) & \textbf{61.39} & \ssonTS{\textbf{57.43}} & 42.57 & \ssonTS{31.68} & 30.69 & \ssonTS{20.79} & 31.68 & \ssonTS{23.76} & 29.70 & \ssonTS{23.76} & 22.77 & \ssonTS{15.84} & 98.02 \\
Detail (101) & \textbf{83.17} & \ssonTS{\textbf{78.22}} & 56.44 & \ssonTS{50.50} & 41.58 & \ssonTS{35.64} & 42.57 & \ssonTS{36.63} & 37.62 & \ssonTS{27.72} & 15.84 & \ssonTS{10.89} & 97.03 \\
\bottomrule
\end{tabular*}
\endgroup
\vspace{-8pt}
\end{table}

\subsection{Main Results}
\label{sec:exp_main}

\myparagraph{Evaluation protocol.}
Six VLMs receive four original, unannotated images in A, B, C, D order with a shared prompt. Independent calls use no demonstrations or conversation history. Inputs exclude camera parameters, top-down views, filenames, and edit annotations. Models explain their choice and end with a JSON answer. Unparsable answers receive no credit. Temperature is zero when supported and otherwise omitted. The token limit is 96,000 for Claude Fable 5 and 16,384 otherwise.

\myparagraph{Scoring decisions and their evidence.}
Answer accuracy credits the selected view. \emph{True solve} additionally requires factual, sufficient visible evidence. Unsupported correct answers are \emph{Partly guessed}. Ten Claude Opus 5 seats screened correct choices, but every such response received an agent-blind human judgment and full human evidence review. An independent human rereview agreed on all unanimous agent-TS cases (Appendix~\ref{app:grading}). Both metrics use all questions as denominators and concern stated evidence, not hidden reasoning. Ten unexposed participants supplied one choice per item, yielding 833/1,000 correct and 83.3\% accuracy (Appendix~\ref{app:human_baseline}).

\myparagraph{Geometric judgments expose the largest capability gap.}
Humans achieve 83.30\% accuracy versus 39.70\% for GPT-6 Astra Pro (Table~\ref{tab:main_choice}). Other VLMs score below 25\%. Uniform guessing is not a performance floor when valid distractors attract systematic errors (Section~\ref{sec:difficulty}). Astra reaches 92.08\% on swap, 27.20\% on move, and 7.23\% on rotate. Across VLMs, rotation peaks at 9.54\% versus 76.01\% for humans, a 66.47-point gap exposing difficulty judging orientation across viewpoints despite evidence supporting stronger human performance.

\myparagraph{Explanation grading changes the assessment of capability.}
Across six models, 404 of 1,368 correct selections (29.5\%) lack sufficient valid support. Astra Pro scores 34.90\% under True solve (Table~\ref{tab:main_true}). Gemini 3.5 Flash leads rotation selection at 9.54\% but achieves only 1.16\% True solve, below Astra Pro's 5.20\%. Thus, selecting the correct view most often need not mean identifying its contradiction most reliably, making explanatory support a consequential evaluation dimension.

Model variants progress unevenly. Gemini 3.6 Flash improves over 3.5 on move and mirror selection but declines elsewhere. Overall answer accuracy falls from 20.60\% to 18.70\%, while True solve rises from 11.60\% to 11.90\%. Neither aggregate score alone captures these changes.

\subsection{Non-VLM Baselines}
\label{sec:exp_encoders}

We test visual distinctiveness with 17 CLIP~\citep{radford2021clip}, ViT~\citep{dosovitskiy2021vit}, and DINOv2~\citep{oquab2024dinov2} encoders, each selecting the image least similar to the other three by mean cosine similarity. For features $z_1,\ldots,z_4$,
\begin{equation}
s_i = \frac{1}{3}\sum_{j\ne i}\frac{z_i^\top z_j}{\lVert z_i\rVert_2\lVert z_j\rVert_2},
\qquad
\hat{y} = \operatorname*{arg\,min}_{i\in\{1,2,3,4\}} s_i.
\end{equation}
Accuracy ranges from 16.9\% to 21.7\% overall (Table~\ref{tab:vision_encoders}, full results in Appendix Table~\ref{tab:vision_encoders_full}) and 5.0\% to 9.9\% on swap, versus 97.03\% for humans. Global appearance dissimilarity thus fails as a reliable proxy for scene inconsistency, supporting the distinction between visually unusual and spatially contradictory observations.

\begin{table}[h]
\centering
\caption{\textbf{Selected vision encoder results (\%) on all 1,000 questions.} The two highest-overall encoders per family select the view with lowest mean cosine similarity to the others. Bold marks the best score among all 17 encoders. Full results appear in Appendix Table~\ref{tab:vision_encoders_full}.}
\label{tab:vision_encoders}
\begingroup
\small
\setlength{\tabcolsep}{2pt}
\renewcommand{\arraystretch}{1.05}
\begin{tabular*}{\linewidth}{@{\extracolsep{\fill}}llrrrrrrr@{}}
\toprule
Family & Encoder & Overall & Swap & Move & Rotate & Mirror & Delete & Detail \\
\midrule
\textit{CLIP} & RN50 & \textbf{21.7} & 8.9 & 25.2 & \textbf{22.5} & 22.8 & 25.7 & 17.8 \\
 & RN50x64 & \textbf{21.7} & 8.9 & \textbf{30.8} & 16.2 & 22.8 & 29.7 & \textbf{21.8} \\
\textit{Vision Transformer} & ViT-L/32 & 21.2 & 6.9 & 29.2 & 17.9 & 21.8 & 27.7 & 19.8 \\
 & ViT-L/16 & 21.1 & 6.9 & 25.2 & 20.5 & 21.8 & 28.7 & 18.8 \\
\textit{DINOv2} & ViT-S/14 & 18.4 & 5.9 & 25.6 & 16.2 & 18.8 & 21.8 & 16.8 \\
 & ViT-g/14 & 18.6 & 6.9 & 23.2 & 17.1 & 20.8 & 24.8 & 15.8 \\
\midrule
\multicolumn{2}{l}{Random guessing} & 25.0 & 25.0 & 25.0 & 25.0 & 25.0 & 25.0 & 25.0 \\
\bottomrule
\end{tabular*}
\endgroup
\vspace{-8pt}
\end{table}

\subsection{Prompting and Camera Information}
\label{sec:exp_prompting}

We test prompting and camera-information interventions using GPT-6 Astra Pro through OpenRouter on 100 fixed diagnostic questions (Table~\ref{tab:intervention_choice}). Selection fixed exactly 12 original True solve and 12 original Partly guessed cases while balancing their counts within each edit category and covering all sources. Base reuses those original responses and has 24\% accuracy and 12\% True solve. Other conditions are new runs, so their differences from Base describe this selected set rather than the full bank. Appendices~\ref{app:intervention_selection} and~\ref{app:prompts} give selection and prompts.

\myparagraph{Separating strategy, task knowledge, and geometry.}
Comp guides comparison through landmarks, instance correspondence, expected visibility, and corroborating views. Scene adds to Comp a requirement to describe the 3D scene arrangement before deciding. Tax supplies the same six-category edit taxonomy for every question without revealing the current edit. Comp+Tax combines Tax and Comp. Cam adds renderer-provided camera intrinsics, extrinsics, and image dimensions to Comp, supplying geometric supervision instead of requiring camera inference. These prompting conditions use one response without reconstruction tools. Comp, Scene, Tax, and Comp+Tax receive no additional scene measurements.

\begin{table}[H]
\centering
\caption{\textbf{Answer accuracy (Acc) and True solve (TS, green) on the 100-question intervention set (\%).} Both use all category questions as denominators. Bold marks the best score without supplied camera calibration (Cam).}
\label{tab:intervention_choice}
\label{tab:intervention_true_solve}
\begingroup
\small
\setlength{\tabcolsep}{0.6pt}
\renewcommand{\arraystretch}{1.05}
\begin{tabular*}{\linewidth}{@{\extracolsep{\fill}}l*{7}{|rr}@{}}
\toprule
Category ($n$) & \multicolumn{2}{c|}{Base} & \multicolumn{2}{c|}{Comp} & \multicolumn{2}{c|}{Scene} & \multicolumn{2}{c|}{Tax} & \multicolumn{2}{c|}{Comp+Tax} & \multicolumn{2}{c|}{Recon} & \multicolumn{2}{c}{Cam} \\
 & Acc & \ssonTS{TS} & Acc & \ssonTS{TS} & Acc & \ssonTS{TS} & Acc & \ssonTS{TS} & Acc & \ssonTS{TS} & Acc & \ssonTS{TS} & Acc & \ssonTS{TS} \\
\midrule
Overall (100) & 24.00 & \ssonTS{12.00} & 40.00 & \ssonTS{23.00} & 37.00 & \ssonTS{29.00} & 44.00 & \ssonTS{32.00} & \textbf{48.00} & \ssonTS{\textbf{34.00}} & 47.00 & \ssonTS{\textbf{34.00}} & 61.00 & \ssonTS{48.00} \\
Swap (18) & 66.67 & \ssonTS{33.33} & \textbf{77.78} & \ssonTS{55.56} & 72.22 & \ssonTS{\textbf{66.67}} & 72.22 & \ssonTS{61.11} & \textbf{77.78} & \ssonTS{61.11} & \textbf{77.78} & \ssonTS{\textbf{66.67}} & 66.67 & \ssonTS{61.11} \\
Move (20) & 10.00 & \ssonTS{5.00} & 20.00 & \ssonTS{20.00} & 20.00 & \ssonTS{20.00} & \textbf{30.00} & \ssonTS{\textbf{25.00}} & 25.00 & \ssonTS{15.00} & 15.00 & \ssonTS{10.00} & 70.00 & \ssonTS{70.00} \\
Rotate (18) & 11.11 & \ssonTS{5.56} & 16.67 & \ssonTS{0.00} & \textbf{22.22} & \ssonTS{\textbf{11.11}} & 16.67 & \ssonTS{\textbf{11.11}} & \textbf{22.22} & \ssonTS{5.56} & 16.67 & \ssonTS{5.56} & 38.89 & \ssonTS{22.22} \\
Mirror (15) & 13.33 & \ssonTS{6.67} & 40.00 & \ssonTS{13.33} & 33.33 & \ssonTS{20.00} & \textbf{66.67} & \ssonTS{\textbf{53.33}} & \textbf{66.67} & \ssonTS{\textbf{53.33}} & 46.67 & \ssonTS{40.00} & 40.00 & \ssonTS{20.00} \\
Delete (12) & 16.67 & \ssonTS{8.33} & 25.00 & \ssonTS{8.33} & 25.00 & \ssonTS{25.00} & 66.67 & \ssonTS{33.33} & \textbf{75.00} & \ssonTS{\textbf{50.00}} & \textbf{75.00} & \ssonTS{\textbf{50.00}} & 75.00 & \ssonTS{58.33} \\
Detail (17) & 23.53 & \ssonTS{11.76} & 58.82 & \ssonTS{35.29} & 47.06 & \ssonTS{29.41} & 23.53 & \ssonTS{11.76} & 35.29 & \ssonTS{29.41} & \textbf{64.71} & \ssonTS{\textbf{41.18}} & 76.47 & \ssonTS{52.94} \\
\bottomrule
\end{tabular*}
\endgroup
\vspace{-12pt}
\end{table}

\myparagraph{Better selection and better support need not coincide.}
Comp reaches 40\% answer accuracy and 23\% True solve. Scene yields fewer correct choices at 37\% but more supported answers at 29\% under current grades, showing distinct benefits from scene description and comparison guidance. Taxonomy knowledge also helps selectively. Tax reaches 66.67\% answer accuracy on mirror and delete but matches Base's 23.53\% on detail. Comp+Tax reaches 48\% overall and 34\% True solve. Knowing possible changes still leaves substantial difficulty locating and verifying the contradiction.

\myparagraph{Supplied geometry most strongly helps displacement judgments.}
Cam reaches 61\% answer accuracy and 48\% True solve, exceeding Comp by 21 and 25 percentage points. On move, both metrics rise from 20\% to 70\%, with valid support for all 14 correct selections. This is consistent with camera geometry helping distinguish projected from scene displacement. Rotation remains harder at 38.89\% selection and 22.22\% True solve. Every condition without supplied calibration remains below 50\% overall answer accuracy. Even Cam provides valid support for fewer than half of all answers.

\subsection{3D Reconstruction}
\label{sec:exp_reconstruction}

Recon uses GPT-6 Astra at the highest of six available reasoning-effort settings for sparse 3D reconstruction and camera-view inspection without supplied geometry. Humans verify the evidence archive and reconstruction use. Noncompliant runs are repeated, keeping the first compliant answer regardless of correctness. The reconstruction prompt is provided in Appendix~\ref{app:prompt_R}.

On this subset, the complete Recon workflow reaches 47\% accuracy and 34\% True solve, exceeding Base by 23 and 22 percentage points. It answers 9/12 delete and 11/17 detail items correctly, versus 3/20 move and 3/18 rotate items. Only two move and one rotate answers qualify as True solve, highlighting persistent difficulty with displacement and orientation.

\subsection{Model Fine-tuning}
\label{sec:exp_finetuning}

We fine-tune Qwen2.5-VL-7B-Instruct~\citep{bai2025qwen25vl} on a grouped, stratified 700/99/201 train/validation/test split. Room and scene-family grouping keeps related items together, with panel SHA-256 leakage screening. Four orders per training question place the intruder at each letter once, yielding 2,800 examples. Headline results use original option order, greedy generation, and a shared final-line JSON parser.

We compare answer supervision with rationale and answer supervision. Generation records supply edit type, object, change, and supporting views only in training targets, remapping view labels per permutation. Both runs train rank-16 language-decoder LoRA~\citep{hu2022lora} adapters for three epochs with the same seed and optimization settings, freezing the visual encoder and merger. Validation selects checkpoints using answer logits or generated responses, respectively. Appendix~\ref{app:finetuning} gives implementation and statistical details.

\begin{table}[H]
\centering
\vspace{-2pt}
\caption{\textbf{Answer accuracy (\%) on 201 held-out questions.} Parentheses give category sizes. Bold marks the best accuracy. Overall intervals are 95\% Wilson confidence intervals.}
\vspace{-2pt}
\label{tab:finetuning}
\begingroup
\small
\setlength{\tabcolsep}{1.2pt}
\renewcommand{\arraystretch}{1.05}
\begin{tabular*}{\linewidth}{@{\extracolsep{\fill}}lrrrrrrrr@{}}
\toprule
Supervision & Overall & Swap & Move & Rotate & Mirror & Delete & Detail & Overall CI \\
 & (201) & (21) & (51) & (67) & (20) & (22) & (20) & (95\%) \\
\midrule
Zero-shot & 21.89 & 14.29 & 23.53 & 23.88 & 25.00 & 18.18 & \textbf{20.00} & [16.7, 28.1] \\
Answer SFT & 33.33 & \textbf{47.62} & \textbf{31.37} & 32.84 & 30.00 & 45.45 & 15.00 & [27.2, 40.1] \\
Rationale + answer SFT & \textbf{41.79} & 38.10 & \textbf{31.37} & \textbf{58.21} & \textbf{35.00} & \textbf{50.00} & 15.00 & [35.2, 48.7] \\
\bottomrule
\end{tabular*}
\endgroup
\vspace{-8pt}
\end{table}

\myparagraph{The task is learnable, with gains concentrated by operation.}
From 21.89\% without fine-tuning, accuracy reaches 33.33\% with answers and 41.79\% with rationales and answers (Table~\ref{tab:finetuning}). All responses parse, ruling out formatting recovery. Both trained models improve over baseline under paired exact McNemar tests. With unchanged test inputs and a frozen visual encoder and merger, the rationale advantage on held-out rooms suggests that supervising scene changes and supporting views can improve attribution beyond answer labels. The additional 8.46-point gain remains suggestive ($p=0.064$, one training seed per condition).

Rotation rises from 32.84\% to 58.21\% with rationales. Its 17 additional correct answers equal the net gain across all 201 test items. Move and detail remain unchanged, while swap declines. Across four cyclic panel orders, mean accuracy rises from 35.3\% with answer supervision to 38.9\% with rationales, and the fraction of questions answered correctly in all four orders rises from 3.0\% to 11.9\%. Most items still lack consistently correct predictions. \bench{} reveals which spatial judgments benefit from supervision and whether gains persist across operations and panel orders.

\section{Conclusions}
\label{sec:conclusion}

\bench{} tests cross-view contradictions through controlled edits and three-stage verification. Humans outperform VLMs, while \emph{True solve} distinguishes supported correct choices. On a selected diagnostic set, prompting and an execution-verified reconstruction workflow attain higher observed scores, with supplied camera geometry highest. Fine-tuning improves accuracy on unseen rooms, particularly for rotation, while order sensitivity remains.
That our findings concern rendered indoor scenes is a limitation of this work. We nonetheless expect the observed trends to hold more broadly, and leave fuller validation to future studies.


\subsection*{AI use statement}
We used generative AI tools to assist with manuscript writing and improve clarity, grammar, and readability. The authors reviewed and revised the AI-assisted text and take full responsibility for the final manuscript, including its claims and citations. VLMs served as experimental systems in the evaluations described in this paper. API evaluations were conducted through OpenRouter. Model responses from the initial development study informed distractor design. The reconstruction condition used GPT-6 Astra at the highest of six available reasoning-effort settings. Ten Claude Opus 5 reviewers assisted with explanation assessment, with every final label determined through human review.

\subsection*{Ethics statement}
\label{sec:ethics}
\bench{} uses rendered views of virtual indoor scenes. The authors consulted the relevant institution about whether human participation in this evaluation task required ethics approval and were advised that it did not. Participants in the human baseline were adults who were informed about the evaluation and participated voluntarily. The ten participants were recruited through the authors' personal networks and were compensated at rates significantly exceeding their local minimum wage. Participants were randomly assigned question packs. Final answer choices were collected for aggregate analysis. No direct personal identifiers or demographic information beyond adult eligibility were collected. The source datasets and simulation environments are credited in Section~\ref{sec:stats}. Findings are specific to this controlled benchmark and do not establish reliability in real-world deployment.

\subsection*{Reproducibility statement}
All 1,000 question sets and the selected fine-tuned checkpoints will be made public. Each question will include a JSON metadata file recording its option camera views, room depth and floor plan, edited-object floor-plan identifier, and edit parameters (Appendix~\ref{app:construction_audit}). Sections~\ref{sec:bench}--\ref{sec:experiments} and the appendices specify construction, review, scoring, baseline, and intervention protocols. Appendix~\ref{app:finetuning} gives the grouped split, leakage checks, optimization, checkpoint selection, and statistical analyses.


\bibliographystyle{iclr2027_conference}

\appendix

\section{Additional Details and Results}
\label{app:details}

\subsection{Fine-tuning Implementation and Additional Diagnostics}
\label{app:finetuning}

\myparagraph{Split construction and input controls.}
The split groups items using declared scene identifiers, scene-family identifiers, and matching panel SHA-256 hashes as leakage-screening relations. Connected items remain within one partition. Stratification uses source and edit category, folding cells with fewer than ten items into a source-specific remainder. Ten shuffled grouped folds allocate two folds to testing, one to validation, and seven to training. Seeds are tried in order against fixed criteria. Each source must contribute at least 30 training, 8 validation, and 12 test items. Source and edit cells with at least ten items require three training and one test item. Each edit requires 30 training and five test items. Test and validation sizes must be 180 to 220 and 80 to 120, with training and test shares differing by at most 0.06 per source and 0.08 per edit. Seeds 0 and 1 fail SAGE rotation coverage in the test set, and seed 2 produces the retained 700/99/201 split. No model predictions or correctness fields determine the split. Original test answer counts are 45, 59, 52, and 45 for A, B, C, and D. A majority-letter predictor therefore scores 29.4\%.

Training augmentation places the intruder at each answer position once and shuffles the three anchors, giving 700 targets per letter. Validation uses one cyclic order per item with approximately balanced answer letters. Main test results use the original order. An additional diagnostic evaluates all four cyclic orders of each test item, preserving image content and mapping predictions to their original panels. The standard image processor handles the full panels, with no manual cropping or image augmentation. Question sheets, answer-key images, and scene metadata are not model inputs.

\myparagraph{Supervision and optimization.}
Rationale targets are deterministically constructed from the recorded edit type, object description, change description, and reference reason. The target describes the altered view and supporting anchors, followed by the answer JSON. View labels are remapped with each panel permutation. Supervision applies token-level cross-entropy to the assistant response only. Prompt and image tokens are masked, and rationale and answer tokens receive equal weight. Distractor identities are excluded from targets.

LoRA adapts all seven linear projections in each of the 28 language-decoder layers, covering 196 modules and 40.4 million trainable parameters. Rank, scaling parameter, and dropout are 16, 32, and 0.05. The visual encoder, visual merger, embeddings, output head, and base weights remain frozen. Both runs use training seed 1, bf16 precision, AdamW~\citep{loshchilov2019adamw}, peak learning rate $10^{-4}$, effective batch size 16, and three epochs comprising 525 steps. The first 16 steps provide linear warm-up, followed by cosine decay. Weight decay is zero and gradient clipping is 1.0. Training uses one H200 with gradient checkpointing. The rationale run resumes from step 464 with optimizer, scheduler, and random-state information preserved.

\myparagraph{Checkpoint selection and evaluation.}
Validation runs every 25 optimizer steps and keeps the most accurate checkpoint, breaking ties in favor of the earlier step. Answer supervision is scored by the four option logits after a fixed JSON prefix and selects step 250. Rationale supervision uses greedy generation capped at 192 new tokens and selects step 425. These scorers match the distinct response formats. Only the selected checkpoints are evaluated on the test set. Both adapters and the base model use greedy test generation with a 16,384-token limit and strict final-line JSON parsing. Every original-order test response parses successfully and none is truncated.

\myparagraph{Uncertainty and order sensitivity.}
Main intervals use the Wilson method over 201 items. Two-sided exact McNemar tests give $p=0.015$ for answer supervision versus the base model and $p=7.9\times10^{-5}$ for rationale supervision versus the base model. Between trained models, 46 items are correct only with rationale supervision and 29 only with answer supervision, giving $p=0.064$. Across cyclic orders, item-level bootstrap intervals from 2,000 resamples are 31.7\% to 39.2\% for answer supervision and 34.7\% to 43.7\% for rationale supervision. The difference is 3.6 percentage points with an interval from $-0.9$ to $8.1$. Resampling items preserves dependence between orders of the same question. The results establish improvements over the base model in this split, while additional seeds are needed to assess the stability of the difference between supervision targets.

\subsection{Complete Vision Encoder Results}
\label{app:vision_encoders}

Table~\ref{tab:vision_encoders_full} reports all 17 encoders evaluated with the appearance-similarity rule in Section~\ref{sec:exp_encoders}.

\begin{table}[!htbp]
\centering
\vspace{-4pt}
\caption{\textbf{Vision encoder answer accuracy (\%) on all 1,000 questions.} Every encoder selects the view with the lowest mean cosine similarity to the other three. Bold marks the best encoder score in each column.}
\label{tab:vision_encoders_full}
\begingroup
\small
\setlength{\tabcolsep}{2pt}
\renewcommand{\arraystretch}{1.12}
\begin{tabular*}{\linewidth}{@{\extracolsep{\fill}}llrrrrrrr@{}}
\toprule
Family & Encoder & Overall & Swap & Move & Rotate & Mirror & Delete & Detail \\
\midrule
\textit{CLIP} & ViT-B/32 & 21.5 & 5.9 & 28.0 & 19.4 & 23.8 & 27.7 & 19.8 \\
 & ViT-B/16 & 19.1 & 5.0 & 27.2 & 15.3 & \textbf{25.7} & 22.8 & 15.8 \\
 & ViT-L/14 & 19.0 & \textbf{9.9} & 27.2 & 13.9 & \textbf{25.7} & 22.8 & 14.9 \\
 & ViT-L/14@336px & 20.0 & \textbf{9.9} & 28.4 & 17.1 & 20.8 & 25.7 & 12.9 \\
 & RN50 & \textbf{21.7} & 8.9 & 25.2 & \textbf{22.5} & 22.8 & 25.7 & 17.8 \\
 & RN101 & 19.8 & 8.9 & 26.0 & 16.2 & 23.8 & \textbf{30.7} & 12.9 \\
 & RN50x4 & 19.3 & 8.9 & 24.8 & 16.8 & 20.8 & 27.7 & 14.9 \\
 & RN50x16 & 21.5 & 6.9 & 24.8 & 21.7 & 22.8 & 27.7 & 19.8 \\
 & RN50x64 & \textbf{21.7} & 8.9 & \textbf{30.8} & 16.2 & 22.8 & 29.7 & \textbf{21.8} \\
\midrule
\textit{Vision Transformer} & ViT-B/32 & 20.8 & 7.9 & 24.4 & 19.4 & 22.8 & \textbf{30.7} & 17.8 \\
 & ViT-B/16 & 19.6 & \textbf{9.9} & 24.8 & 15.3 & 21.8 & 28.7 & 19.8 \\
 & ViT-L/32 & 21.2 & 6.9 & 29.2 & 17.9 & 21.8 & 27.7 & 19.8 \\
 & ViT-L/16 & 21.1 & 6.9 & 25.2 & 20.5 & 21.8 & 28.7 & 18.8 \\
\midrule
\textit{DINOv2} & ViT-S/14 & 18.4 & 5.9 & 25.6 & 16.2 & 18.8 & 21.8 & 16.8 \\
 & ViT-B/14 & 16.9 & 5.9 & 22.0 & 11.8 & 20.8 & 24.8 & 20.8 \\
 & ViT-L/14 & 17.8 & 8.9 & 23.2 & 17.6 & 17.8 & 17.8 & 13.9 \\
 & ViT-g/14 & 18.6 & 6.9 & 23.2 & 17.1 & 20.8 & 24.8 & 15.8 \\
\midrule
\multicolumn{2}{l}{Random guessing} & 25.0 & 25.0 & 25.0 & 25.0 & 25.0 & 25.0 & 25.0 \\
\bottomrule
\end{tabular*}
\endgroup
\vspace{-6pt}
\end{table}

\subsection{Model Sources and Checkpoint}
\label{app:model_sources}

Table~\ref{tab:model_sources} links the evaluated model names to their OpenRouter model pages. Fine-tuning uses the checkpoint \texttt{Qwen/Qwen2.5-VL-7B-Instruct}, available from the \href{https://huggingface.co/Qwen/Qwen2.5-VL-7B-Instruct}{Qwen model repository}.

\begin{table}[!htbp]
\centering
\caption{\textbf{Reference pages for the evaluated VLMs.} OpenRouter links identify the models. Experimental settings are reported separately. Links last checked on September 25, 2026.}
\label{tab:model_sources}
\begin{tabular}{@{}ll@{}}
\toprule
Model & Model reference page \\
\midrule
GPT-6 Astra Pro & \href{https://openrouter.ai/openai/gpt-6-astra-pro}{OpenRouter model page} \\
GPT-5.6 Sol Pro & \href{https://openrouter.ai/openai/gpt-5.6-sol-pro}{OpenRouter model page} \\
Gemini 3.5 Flash & \href{https://openrouter.ai/google/gemini-3.5-flash}{OpenRouter model page} \\
Claude Fable 5 & \href{https://openrouter.ai/anthropic/claude-fable-5}{OpenRouter model page} \\
Gemini 3.6 Flash & \href{https://openrouter.ai/google/gemini-3.6-flash}{OpenRouter model page} \\
Grok 4.5 & \href{https://openrouter.ai/x-ai/grok-4.5}{OpenRouter model page} \\
\bottomrule
\end{tabular}
\end{table}

\subsection{Construction Audit and Item Metadata}
\label{app:construction_audit}

The final bank was selected from more than 10,000 development candidates. The three human checks in Section~\ref{sec:human_verification} preceded all benchmark evaluations. The first reviewer could inspect A--D, the answer key, and paired top-down views. They checked the edit against the recorded scene, physical and rendering integrity, and whether its diagnostic features were usable in the intruder and at least two anchors. The remaining anchor needed environmental evidence linking it to the same room. Interpenetration, floating objects, unsupported placement, clipped or distorted diagnostic objects, and evidence erased by framing or occlusion caused rejection. For a deletion, the expected location had to be visible from the intruder camera so that absence could be distinguished from occlusion.

In the second check, a reviewer who had not seen the question answered from the four option images alone. They recorded their selected option, visual basis, and any ambiguity before seeing the key. The reviewer then compared the key, investigated any mismatch, and checked whether another option could also be the intruder even when their first answer matched. Suspected alternative answers were checked with the first-stage reviewer. Confirmed ambiguity or item error caused rejection. In the third check, reviewers considered whether the item could be solved using isolated object detection, an object inventory, an isolated appearance difference, or nearly aligned pixel comparisons without the intended cross-view attribution. Items admitting those tested shortcuts were rejected. This audit targets specified shortcut families rather than claiming to rule out every possible heuristic. The 1,000 accepted questions were frozen after these checks. The later model-explanation defect check led to no deletion or replacement.

Automated screening used an edited-region threshold of at least 0.25\% of intruder image pixels and a target-footprint gate of 0.4--16\% of image area. Reviewers also rejected targets blending into a frame edge and required clear diagnostic evidence and scene references. The area criteria by themselves do not establish answerability. The planned release includes A--D for each item and a corresponding \texttt{repro.json} with the four camera views, room depth, floor plan, edited object's floor-plan number, and edit parameters. Per-item metadata permits inspection of the recorded edit without supplying these fields as evaluation inputs.

\subsection{Human Answer Baseline}
\label{app:human_baseline}

Ten adult participants who had not designed the questions and had no advance knowledge of the construction procedure completed the benchmark through a web interface backed by Supabase and Vercel. The 1,000 questions were shuffled into 50 disjoint packs of 20. Each participant was randomly assigned five packs. The interface showed the four option images and asked which view was inconsistent. Each question received one final option judgment, for 1,000 judgments in all. The back-end aggregate was cross-checked against pack totals. Exactly 833 final choices matched the key, so human answer accuracy is $833/1000=83.3\%$. This is an aggregate of single judgments, not a per-item majority vote or a measurement of agreement between people. Only final choices, without identifying participant data or response processes, entered the analysis.

\subsection{True Solve Grading}
\label{app:grading}

Across six VLMs, 1,368 of the 6,000 original answers select the correct view. Of these, 964 have sufficient stated evidence and 404 do not. The latter group includes fabricated changes and explanations that are too vague or otherwise insufficient. It must not be interpreted as 404 observed hallucinations. Grading uses the complete visible answer, not encrypted or inferred reasoning. An answer that identifies a real scene change, even incompletely or without the generator's edit-category name, is \emph{True solve} when the claimed evidence suffices to justify the selected view. A fabricated asserted change, a viewpoint effect presented as an edit, or insufficient evidence precludes True solve. Hypotheses that the solver explicitly tests and rejects are distinguished from claims it adopts as facts. The same grading protocol was applied to correct-choice responses in all intervention conditions.

\myparagraph{Agent screening.}
Ten separate reviewer seats used Claude Opus 5 at maximum reasoning effort. Four seats independently reviewed every correct-choice response and voted \texttt{TS}, \texttt{HALLUCINATION}, or \texttt{DISPUTE}. Only an initial unanimous TS or unanimous HALLUCINATION proceeded to three stricter rounds, with two new seats in each round. Every later seat reviewed every forwarded response. A disagreement or DISPUTE vote at any point routed the response to the agent-dispute class. Only unanimous votes across all ten seats yielded an agent TS or agent HALLUCINATION class. Thus ten votes were available for a forwarded response, not for every correct-choice response. The agent HALLUCINATION vote means it found a fabricated change without identifying the real edit. It does not mean that every final correct-but-unsupported answer contained a hallucination.

Each agent received the original four images, a trusted edit-fact record with the answer, edited object and approximate image locations, and the complete visible correct-choice reply with its parsed final option. Solver identities and prompting conditions were hidden, as were other seats' and rounds' judgments. Packets and replies were shuffled separately for each seat. The reviewers did not see hidden solver reasoning or answer-key images. They localized the edit and response claims in the four images, distinguished asserted facts from tentative hypotheses, and recorded evidence for each vote. A script checked packet coverage, allowed fields and model identifiers, exact quotations from the response, the required image locations for false claims, and logical agreement of the vote with the recorded answers to three rubric questions. Failed format checks were returned to the same seat for format correction only. The checker did not determine visual truth. The reviewer instructions and round headers appear in Appendix~\ref{app:grader_prompt}.

\myparagraph{Human decision.}
Every correct-choice response, including both unanimous agent classes and all agent disputes, went to a human reviewer. The reviewer first examined the four images and the complete visible response without seeing the agent classification, made an independent preliminary judgment, then inspected the agents' localized edit and response evidence and reviewed every record again. The final label was the human decision even if it differed from the agent vote. An additional human independently reviewed the unanimously agent-TS responses and agreed with the TS classification in every case, giving 100\% agreement for that subset. A separate human checked any response flagged as hallucinated by an agent or the first human against A--D for a rendering or question defect. This defect check found no reason to change the frozen bank. This conditional agreement does not estimate agreement across all final labels.

\myparagraph{Wording and factual boundaries.}
High-similarity object names and sufficient partial descriptions count. ``The cup on the table'' can identify an edited white cup without specifying its color, provided the answer does not falsely identify the unchanged black cup. For a swap, correctly locating one object's displacement or describing its disappearance from a particular place and the other object's appearance \emph{at that place} can suffice. Saying it vanished from the room when it remains elsewhere cannot. Generic ``moved'' can describe an actual rotation, but an explicit nonexistent translation cannot. A deletion may be described as moving an object \emph{away} from the scene. A plain relocation claim does not establish deletion. A substitution can be described by the old object leaving and a new one appearing at its location, whereas recoloring does not license an asserted entity deletion. A mirror answer must identify reflection of the room or an existing part of it. A correct letter with invented changes, no assessable explanation, or vague evidence remains correct by answer accuracy but is not True solve. These examples illustrate the fixed rubric and are not quotations from model answers.

\subsection{Intervention Subset and Response Accounting}
\label{app:intervention_selection}

The diagnostic set draws 100 questions from the frozen 1,000-question bank using Astra Pro's original answer and True solve outcomes. Selection used explicit outcome quotas together with edit-category and source-coverage constraints. The conditions required exactly 12 original True solve and 12 original correct-but-unsupported responses. Within each of the six edit categories, the selected counts of these two outcomes matched. The other 76 Base responses were wrong. Each edit category contributed at least 12 questions, and all seven sources were represented. The same 100 question identifiers were fixed for all intervention conditions. Base directly reuses the main evaluation's original responses and labels. It is not an independently repeated baseline run. Selection conditions on that model's observed outcomes, so the results describe a diagnostic set rather than a population-average estimate over SSoN. Table~\ref{tab:intervention_counts} gives exact outcome counts for every condition.

\begin{table}[!htbp]
\centering
\caption{\textbf{Correct-choice and True solve counts on the fixed diagnostic set.} These counts correspond to the percentages in Table~\ref{tab:intervention_choice}. They are category-level outcome totals. They do not identify how individual questions changed status relative to Base.}
\label{tab:intervention_counts}
\begingroup
\small
\setlength{\tabcolsep}{5pt}
\begin{tabular}{@{}lrrrrrrr@{}}
\toprule
Correct choices & Base & Comp & Scene & Tax & Comp+Tax & Recon & Cam \\
\midrule
Swap & 12 & 14 & 13 & 13 & 14 & 14 & 12 \\
Move & 2 & 4 & 4 & 6 & 5 & 3 & 14 \\
Rotate & 2 & 3 & 4 & 3 & 4 & 3 & 7 \\
Mirror & 2 & 6 & 5 & 10 & 10 & 7 & 6 \\
Delete & 2 & 3 & 3 & 8 & 9 & 9 & 9 \\
Detail & 4 & 10 & 8 & 4 & 6 & 11 & 13 \\
\textbf{Overall} & \textbf{24} & \textbf{40} & \textbf{37} & \textbf{44} & \textbf{48} & \textbf{47} & \textbf{61} \\
\midrule
True solve & Base & Comp & Scene & Tax & Comp+Tax & Recon & Cam \\
\midrule
Swap & 6 & 10 & 12 & 11 & 11 & 12 & 11 \\
Move & 1 & 4 & 4 & 5 & 3 & 2 & 14 \\
Rotate & 1 & 0 & 2 & 2 & 1 & 1 & 4 \\
Mirror & 1 & 2 & 3 & 8 & 8 & 6 & 3 \\
Delete & 1 & 1 & 3 & 4 & 6 & 6 & 7 \\
Detail & 2 & 6 & 5 & 2 & 5 & 7 & 9 \\
\textbf{Overall} & \textbf{12} & \textbf{23} & \textbf{29} & \textbf{32} & \textbf{34} & \textbf{34} & \textbf{48} \\
\bottomrule
\end{tabular}
\endgroup
\end{table}

The overall numbers of correct but unsupported responses are 12 for Base, 17 for Comp, 8 for Scene, 12 for Tax, 14 for Comp+Tax, 13 for Recon, and 13 for Cam. The corresponding wrong-choice counts are 76, 60, 63, 56, 52, 53, and 39. These marginal totals do not specify item-level transitions relative to Base.

\subsection{Reconstruction Compliance and Interpretation}
\label{app:reconstruction_audit}

The Recon prompt in Appendix~\ref{app:prompt_R} allowed an initial reconstruction-and-inspection attempt and, if its results justified it, at most one correction \emph{within a run}. Separately, the study reran a whole item when the response lacked a \texttt{reconstruction\_evidence.zip} or did not actually execute and use a sparse 3D reconstruction. A human inspected the submitted archive for an executed spatial representation and reconstructed camera views, and checked the answer's visible account for genuine use of that reconstruction in reasoning. A point cloud or mesh was not required. A merely proposed program or an unsupported claim of reconstruction did not qualify. The first response meeting these execution and evidence conditions was retained whether its selected letter was right or wrong. All 100 scored Recon responses therefore satisfied the execution criterion. The reported 47 correct choices and 34 True solve responses measure task outcomes under this execution-verified reconstruction workflow. External reruns affect compute and response selection and must not be conflated with the prompt's within-run correction limit. Comparisons with Base evaluate the complete reconstruction workflow, including differences in model configuration, reasoning effort, tool access, instructions, and compliance reruns.

\section{Prompts}
\label{app:prompts}

This appendix records the instructions used to test whether comparison guidance, edit-taxonomy knowledge, supplied camera geometry, or reconstruction tools improve performance beyond the original task prompt. Full prompts and accompanying input descriptions make the information available in each condition explicit, separating changes in instructions from additional geometric inputs. Line wrapping and punctuation are normalized for presentation, and Recon wording is edited for clarity. Image content and item-specific camera data are separate inputs and are not part of the fixed prompt text.

\subsection{Original task prompt (Base)}
\label{app:prompt_B}

The original task instruction is reproduced below. The four option images
are supplied in A, B, C, D order.

\begin{quote}
\small\fontencoding{T1}\ttfamily\raggedright
You will be shown four images labeled A, B, C, and D (in that order). Each image shows a scene from one camera viewpoint.\par
\medskip
Three of the four images are mutually consistent views of one and the same scene, taken from different camera positions and angles. One image is inconsistent with that scene. It cannot be a view of the same scene as the other three. Identify which one.\par
\medskip
Reason step by step, then end your response with ONLY this JSON on the final line:\par
\{"odd\_one\_out": "\textless{}A\textbar{}B\textbar{}C\textbar{}D\textgreater{}"\}\par
\end{quote}

\subsection{Evidence comparison strategy (Comp)}
\label{app:prompt_P}

Comp adds a comparison strategy to the original task instruction.
The complete prompt is reproduced below.

\begin{quote}
\small\fontencoding{T1}\ttfamily\raggedright
You will be shown four images labeled A, B, C, and D (in that order). Each image shows a scene from one camera viewpoint.\par
\medskip
Three of the four images are mutually consistent views of one and the same scene, taken from different camera positions and angles. One image is inconsistent with that scene. It cannot be a view of the same scene as the other three. Identify which one.\par
\medskip
Use the following comparison strategy:\par
1. Establish a shared scene reference using stable room landmarks and surrounding objects. Image-left/right, projected position, and apparent facing direction can change with the camera viewpoint.\par
2. Match object instances using their location and surrounding context, especially when several objects look alike. Check the object's boundary before assigning a feature to it.\par
3. Distinguish a real scene change from viewpoint-dependent appearance, occlusion, or field of view. An object that appears only in a newly revealed region is not by itself a contradiction. Absence is informative only when the relevant location is observable and the object should be visible there.\par
4. When two views seem to conflict, use additional views to determine which configuration is corroborated. Partial observations can provide complementary evidence; a view with insufficient overlap is not automatically inconsistent.\par
5. Check the proposed exclusion: the other three views should remain compatible, and the excluded view should have a specific contradiction with their shared scene. Identify which object or property changed using stable surroundings, rather than assigning a change to whichever object is easiest to name.\par
Base your explanation on a concrete, sufficient visual discrepancy. Separate observed evidence from uncertain inference, and do not invent objects, measurements, orientations, or visibility that the images do not establish.\par
\medskip
Reason step by step, then end your response with ONLY this JSON on the final line:\par
\{"odd\_one\_out": "\textless{}A\textbar{}B\textbar{}C\textbar{}D\textgreater{}"\}\par
\end{quote}

\subsection{Qualitative scene description (Scene)}
\label{app:prompt_S}

Scene retains Comp and asks the model to generate a qualitative description of the
scene before making its decision. This description is part of the model
response, not supplied ground-truth scene information.

\begin{quote}
\small\fontencoding{T1}\ttfamily\raggedright
You will be shown four images labeled A, B, C, and D (in that order). Each image shows a scene from one camera viewpoint.\par
\medskip
Three of the four images are mutually consistent views of one and the same scene, taken from different camera positions and angles. One image is inconsistent with that scene. It cannot be a view of the same scene as the other three. Identify which one.\par
\medskip
Use the following comparison strategy:\par
1. Establish a shared scene reference using stable room landmarks and surrounding objects. Image-left/right, projected position, and apparent facing direction can change with the camera viewpoint.\par
2. Match object instances using their location and surrounding context, especially when several objects look alike. Check the object's boundary before assigning a feature to it.\par
3. Distinguish a real scene change from viewpoint-dependent appearance, occlusion, or field of view. An object that appears only in a newly revealed region is not by itself a contradiction. Absence is informative only when the relevant location is observable and the object should be visible there.\par
4. When two views seem to conflict, use additional views to determine which configuration is corroborated. Partial observations can provide complementary evidence; a view with insufficient overlap is not automatically inconsistent.\par
5. Check the proposed exclusion: the other three views should remain compatible, and the excluded view should have a specific contradiction with their shared scene. Identify which object or property changed using stable surroundings, rather than assigning a change to whichever object is easiest to name.\par
Base your explanation on a concrete, sufficient visual discrepancy. Separate observed evidence from uncertain inference, and do not invent objects, measurements, orientations, or visibility that the images do not establish.\par
\medskip
Before deciding, construct a compact qualitative 3D scene description in your written response. Use only the landmarks and objects needed for the decision. Describe their scene-relative arrangement, relevant facing directions, and which views support or leave each relationship uncertain. Then use this description to check the candidate contradiction. Do not invent metric coordinates or fill unseen regions. No reconstruction program or external geometry tool is available in this condition.\par
\medskip
Reason step by step, then end your response with ONLY this JSON on the final line:\par
\{"odd\_one\_out": "\textless{}A\textbar{}B\textbar{}C\textbar{}D\textgreater{}"\}\par
\end{quote}

\subsection{Supplied camera calibration (Cam)}
\label{app:prompt_C}

Cam retains Comp and supplies ground-truth camera calibration as a separate input
after the four images. The fixed instruction is reproduced below. The per-item
numeric calibration is accompanied by a description of the coordinate
convention and image dimensions.

\begin{quote}
\small\fontencoding{T1}\ttfamily\raggedright
You will be shown four images labeled A, B, C, and D (in that order). Each image shows a scene from one camera viewpoint.\par
\medskip
Three of the four images are mutually consistent views of one and the same scene, taken from different camera positions and angles. One image is inconsistent with that scene. It cannot be a view of the same scene as the other three. Identify which one.\par
\medskip
Use the following comparison strategy:\par
1. Establish a shared scene reference using stable room landmarks and surrounding objects. Image-left/right, projected position, and apparent facing direction can change with the camera viewpoint.\par
2. Match object instances using their location and surrounding context, especially when several objects look alike. Check the object's boundary before assigning a feature to it.\par
3. Distinguish a real scene change from viewpoint-dependent appearance, occlusion, or field of view. An object that appears only in a newly revealed region is not by itself a contradiction. Absence is informative only when the relevant location is observable and the object should be visible there.\par
4. When two views seem to conflict, use additional views to determine which configuration is corroborated. Partial observations can provide complementary evidence; a view with insufficient overlap is not automatically inconsistent.\par
5. Check the proposed exclusion: the other three views should remain compatible, and the excluded view should have a specific contradiction with their shared scene. Identify which object or property changed using stable surroundings, rather than assigning a change to whichever object is easiest to name.\par
Base your explanation on a concrete, sufficient visual discrepancy. Separate observed evidence from uncertain inference, and do not invent objects, measurements, orientations, or visibility that the images do not establish.\par
\medskip
After the four original option images, you will receive the renderer's camera calibration for A, B, C, and D, together with its coordinate convention and image dimensions. This is additional ground-truth camera information. It does not identify a trusted view or changed object. Use it to interpret viewpoint differences, then resolve the inconsistency from the images. Camera differences alone are not evidence that a view is inconsistent.\par
\medskip
Reason step by step, then end your response with ONLY this JSON on the final line:\par
\{"odd\_one\_out": "\textless{}A\textbar{}B\textbar{}C\textbar{}D\textgreater{}"\}\par
\end{quote}

The archived camera-input wrapper is reproduced below. Item-specific numeric JSON packets are separate evaluation inputs and are not reproduced here.

\begin{quote}
\small\fontencoding{T1}\ttfamily\raggedright
The four original option images have ended. Below are the numeric renderer camera calibrations for these exact images. The numeric camera data use a perspective pinhole model. width and height are the original image dimensions in pixels. K is the 3-by-3 intrinsic matrix in those pixels. Camera axes are x right, y down, z forward. T\_reference\_to\_camera maps coordinates in camera A's reference frame to each view's camera frame using a 4-by-4 homogeneous matrix. Camera A defines the coordinate frame only and is not designated as trustworthy. All camera translations share one scale, normalized by the median nonzero pairwise camera-center distance; when all centers coincide, translations are zero. Camera differences alone are not evidence of a scene change. No scene state, object metadata, or answer is supplied.\par
\medskip
Camera calibration:\par
\end{quote}

\subsection{Edit taxonomy (Tax)}
\label{app:prompt_T}

Tax adds the same six-category edit taxonomy to every question.
It does not reveal the edit category of the current question.

\begin{quote}
\small\fontencoding{T1}\ttfamily\raggedright
You will be shown four images labeled A, B, C, and D (in that order). Each image shows a scene from one camera viewpoint.\par
\medskip
Three of the four images are mutually consistent views of one and the same scene, taken from different camera positions and angles. One image is inconsistent with that scene. It cannot be a view of the same scene as the other three. Identify which one.\par
\medskip
In this benchmark, three views are rendered from an unchanged scene, while one view is rendered from a version of that scene with a controlled edit. The edit belongs to one of these six categories:\par
- Swap: Two objects exchange locations; their orientations may also be exchanged.\par
- Move: An object's position in the scene changes.\par
- Rotate: An object's orientation in the scene changes.\par
- Mirror: The entire room is reflected left-right.\par
- Delete: An object present in the unchanged scene is removed.\par
- Detail: An object's shape or appearance changes.\par
This list specifies the possible edit categories for the benchmark. It does not specify which category applies to the current question.\par
\medskip
Reason step by step, then end your response with ONLY this JSON on the final line:\par
\{"odd\_one\_out": "\textless{}A\textbar{}B\textbar{}C\textbar{}D\textgreater{}"\}\par
\end{quote}

\subsection{Comparison strategy and edit taxonomy (Comp+Tax)}
\label{app:prompt_PT}

Comp+Tax combines the complete comparison strategy from Comp with the same fixed
taxonomy used in Tax.

\begin{quote}
\small\fontencoding{T1}\ttfamily\raggedright
You will be shown four images labeled A, B, C, and D (in that order). Each image shows a scene from one camera viewpoint.\par
\medskip
Three of the four images are mutually consistent views of one and the same scene, taken from different camera positions and angles. One image is inconsistent with that scene. It cannot be a view of the same scene as the other three. Identify which one.\par
\medskip
Use the following comparison strategy:\par
1. Establish a shared scene reference using stable room landmarks and surrounding objects. Image-left/right, projected position, and apparent facing direction can change with the camera viewpoint.\par
2. Match object instances using their location and surrounding context, especially when several objects look alike. Check the object's boundary before assigning a feature to it.\par
3. Distinguish a real scene change from viewpoint-dependent appearance, occlusion, or field of view. An object that appears only in a newly revealed region is not by itself a contradiction. Absence is informative only when the relevant location is observable and the object should be visible there.\par
4. When two views seem to conflict, use additional views to determine which configuration is corroborated. Partial observations can provide complementary evidence; a view with insufficient overlap is not automatically inconsistent.\par
5. Check the proposed exclusion: the other three views should remain compatible, and the excluded view should have a specific contradiction with their shared scene. Identify which object or property changed using stable surroundings, rather than assigning a change to whichever object is easiest to name.\par
Base your explanation on a concrete, sufficient visual discrepancy. Separate observed evidence from uncertain inference, and do not invent objects, measurements, orientations, or visibility that the images do not establish.\par
\medskip
In this benchmark, three views are rendered from an unchanged scene, while one view is rendered from a version of that scene with a controlled edit. The edit belongs to one of these six categories:\par
- Swap: Two objects exchange locations; their orientations may also be exchanged.\par
- Move: An object's position in the scene changes.\par
- Rotate: An object's orientation in the scene changes.\par
- Mirror: The entire room is reflected left-right.\par
- Delete: An object present in the unchanged scene is removed.\par
- Detail: An object's shape or appearance changes.\par
This list specifies the possible edit categories for the benchmark. It does not specify which category applies to the current question.\par
\medskip
Reason step by step, then end your response with ONLY this JSON on the final line:\par
\{"odd\_one\_out": "\textless{}A\textbar{}B\textbar{}C\textbar{}D\textgreater{}"\}\par
\end{quote}

\subsection{Tool-based reconstruction (Recon)}
\label{app:prompt_R}

The 100-question Recon condition used GPT-6 Astra at the highest of six available reasoning-effort settings. The reconstruction instructions below are edited for clarity.

\begin{quote}
\small\fontencoding{T1}\ttfamily\raggedright
You will be shown four images labeled A, B, C, and D (in that order). Each image shows a scene from one camera viewpoint.\par
\medskip
Three of the four images are mutually consistent views of one and the same scene, taken from different camera positions and angles. One image is inconsistent with that scene. It cannot be a view of the same scene as the other three. Identify which one.\par
\medskip
Use the available execution tools to attempt an explicit three-dimensional reconstruction from these images and inspect whether the resulting geometry supports your answer. You must choose and carry out the method yourself. The original image files are available under the anonymous names A.png, B.png, C.png, and D.png. No ground-truth camera parameters or additional scene observations are provided.\par
\medskip
Tool environment: use only the execution tools provided for this run, their available general-purpose libraries, and programs you write yourself. Do not use external compute, external model APIs, downloaded pretrained reconstruction models, network search, or additional package installations. Work within the capabilities that are actually available.\par
\medskip
First perform one reconstruction-and-inspection attempt. If its results justify a correction, you may perform at most one further reconstruction-and-inspection attempt. Do not start an open-ended search or repeated restarts. A coarse or partial three-dimensional representation is acceptable; clearly state its extent, assumptions, and limitations. Save the code, any geometric representation you actually produce, and the outputs of your checks. Distinguish estimated geometry from observed evidence. Do not claim that unexecuted code or an imagined reconstruction was executed.\par
\medskip
After these attempts, answer using the evidence you actually obtained. If reconstruction fails or remains inconclusive, report that limitation and still provide your best supported judgment. Explain the visual discrepancy supporting your choice and what the executed geometry checks did or did not establish. Then end your response with ONLY this JSON on the final line:\par
\{"odd\_one\_out": "\textless{}A\textbar{}B\textbar{}C\textbar{}D\textgreater{}"\}\par
\end{quote}

\subsection{Fine-tuning prompt and supervision targets}
\label{app:prompt_finetuning}

The Qwen2.5-VL-7B zero-shot evaluation and both fine-tuning conditions use the
same user instruction shown in Appendix~\ref{app:prompt_B}. The task prompt is
followed by the four images in their current A, B, C, D order. The caller supplies
no additional system message. The model's chat template inserts its default
system message, reproduced below, during both training and evaluation.

\begin{quote}
\small\fontencoding{T1}\ttfamily\raggedright
You are a helpful assistant.\par
\end{quote}

Answer-only supervision changes the assistant target, not the user prompt. The
following example illustrates the target for a question whose correct option
is B. The terminal token is shown explicitly.

\begin{quote}
\small\fontencoding{T1}\ttfamily\raggedright
\{"odd\_one\_out": "B"\}\textless{}\textbar{}im\_end\textbar{}\textgreater{}\par
\end{quote}

Rationale-and-answer supervision uses the same user prompt with a structured
assistant target. The example below is reproduced from the training report.
Its fields are derived from the recorded edit and its supporting views.

\begin{quote}
\small\fontencoding{T1}\ttfamily\raggedright
Edit type: rotate\par
Edited object: one of the four dining chairs\par
What changed: DiningChair (one of four) rotated about its vertical axis.\par
Reason: In view D, one of the four dining chairs is turned to a different orientation than in views A and C.\par
\{"odd\_one\_out": "D"\}\textless{}\textbar{}im\_end\textbar{}\textgreater{}\par
\end{quote}

The option letters in the target are remapped with each training permutation.
The edit description and rationale are supervised outputs and are not supplied
in the test prompt. Both fine-tuned models and the zero-shot baseline retain
the same final-line JSON requirement.

\subsection{Reviewer Instructions}
\label{app:grader_prompt}

The operational instructions and the first-round and strict-round headers appear below. The text omits repeated author quotations while retaining the operational questions, voting rules, decision boundaries, output schema, and round procedures. Seat identity and assigned packet folders varied by dispatch. Premise P1 describes the keyed construction review and image-only answerability review. The third-stage shortcut audit described in Appendix~\ref{app:construction_audit} also preceded scoring. The operative TS, HALLUCINATION, and DISPUTE votes apply only to correct-choice packets. The embedded general-purpose rubric describes other response types and is not the final routing decision.

\begin{Verbatim}[breaklines=true,breakanywhere=true,breaksymbolleft={},breaksymbolright={},fontsize=\small]
===== Section 3 COMMON PROMPT (used by all ten seats, after the Section 4 round header) =====

# SSoN true-solve review  -  reviewer instructions

## 1. Your role

You are an independent reviewer for SSoN, a benchmark of four-view scene inconsistency. Each item shows one room
from four cameras, A to D. Exactly one view, the intruder, shows the room after a single controlled edit; the other
three show the unedited room. A solver model chose one view and explained its choice. Every reply you receive chose
the correct view. Your job is to decide, from the images, whether the reply actually found the real edit, or reached
the right letter on claims that the images contradict. You are not the solver and you never communicate with it.

## 2. What you may use

Only the files in your packet folder:

- `A.png`, `B.png`, `C.png`, `D.png`: the four original question images.
- `facts.json`: the trusted answer and the facts of the one edit (section 4).
- `replies.json`: the replies. Each has `unit_id`, `parsed_final_choice` (read mechanically from the reply's last
  line) and `visible_response` (the complete visible text of the reply).

You are not given, and must not look for: the solver's hidden reasoning, the solver's identity, its prompting
condition, the answer-key image, other packets, other reviewers' records, earlier rounds, or any other file on this
machine. Do not use outside knowledge of the rendering engine or of other items.

## 3. Premises (binding; set by the study owner)

P1. Every item passed two rounds of human construction review. The first round checked the item against its answer
    key: the edit is present and clearly visible, with sufficient evidence, in the intruder and in at least two other
    views; the remaining view shows the same room; there is no clipping, floating or other physical or rendering
    problem. The second round had fresh reviewers solve the item from the four images alone and confirmed that it has
    one unique answer.

P2. Each item has exactly one edit, the one in `facts.json`. Every other difference between the views (position on
    screen, apparent size, facing direction, occlusion, objects entering or leaving the frame, lighting and shading)
    is caused by the change of viewpoint and is not a scene change.

P3. No image contains a rendering or physics problem. Never explain or excuse a reply's claim by an item error. A
    reply that reports floating, clipping, interpenetration, missing support, glitches or render artefacts is
    reporting a change that does not exist.


## 4. facts.json

- `trusted_answer`: the intruder letter.
- `edit.type`: move, rotate, swap, delete, detail or mirror.
- `edit.objects`: for each edited object, its name and category, a `descriptor` that tells it apart from similar
  objects, and `same_type_company`: other objects of the same category in the views, which were NOT edited.
- `edit.detail_kind` (detail only): `substitution` (the object was replaced by a different object) or `appearance`
  (the same object with a new colour, texture or surface condition).
- `edit.magnitude`: metres for move, degrees for rotate, when recorded.
- `edit.assisting_translation` (rotate only, when present): a small translation that was part of the edit, with its
  size and direction. A reply's translation claim that matches it is true.
- `edit.swap_orientation` (swap only): whether each object kept its own orientation or the orientations were
  exchanged.
- `edit.description`: one sentence stating the edit.
- `edited_object_image_locations`: approximate pixel positions (x right, y down) of each edited object's place in
  every view, projected from its recorded position. Occlusion is not accounted for, so a point can be hidden behind
  something, and a centre just outside the frame can still leave part of the object visible at the edge.

## 5. Fixed grader prompt and label definitions (verbatim from the study instructions)

```text
Evaluate the factual sufficiency of the supplied visible response or trajectory for this four-view inconsistency task. You are an independent reviewer, not the solver. Do not send feedback to the solver.

Use only the original A-D images, trusted answer/edit facts supplied for review, and the visible response/tool-event record. Treat generated reconstructions as predictions, not ground truth. If trusted facts are unavailable or conflicting, explain the limitation and return undetermined with blocked_missing_gt or adjudication_pending. Do not infer correctness from confident wording.

For a completed answer, first check its final choice against the trusted answer. A wrong choice is wrong, never partly_guessed. A correct choice is true_solve only when its stated visual discrepancy is factual and sufficient to support the selection. Valid partial explanations qualify; naming both a complete object identity and an edit taxonomy label is not mandatory. A correct choice supported by fabricated discrepancies, viewpoint effects mistaken for scene edits, or insufficient vague evidence is partly_guessed. A correct choice with no assessable explanation is correct_unassessable. Missing, truncated, or invalid final answers are technical_failure.

For response-level review of an agent turn without a final answer, use intermediate_no_final and still examine any factual scene claims. If the turn contains only tool calls and no scene claims, explicitly state no_scene_claims and its role in the investigation. This is not a guessed answer. At trajectory level, use the final choice and all visible model statements, not just the last sentence. Distinguish tentative hypotheses from asserted facts and raw noisy tool predictions from claims adopted by the solver. A false prediction does not excuse an unsupported factual claim.

Return one JSON object using the supplied field specification. Include concise analysis explaining the verdict and a claims array: short exact response excerpts, image labels and locations supporting or refuting them, and whether each statement is a fact, a tentative hypothesis, or a tool prediction. Never invent a quotation or unseen evidence. Retain externally assigned unit identifiers; do not invent them. Automated review_status is preliminary unless blocked_missing_gt or adjudication_pending is required. Do not mark your own automated review as human-final.
```



How these apply here: sections 6 to 9 are the operational rules of this review and take precedence where they are
more specific. Your vote replaces the label set and maps to it as follows: TS  maps to  true_solve; HALLUCINATION  maps to 
partly_guessed; DISPUTE  maps to  undetermined with adjudication_pending, decided by a human. In particular, a reply that
neither identifies the edit nor asserts any false change (vague, eliminative or unexplained) is DISPUTE here, and a
human assigns its final label. Your review is always preliminary.

## 6. Three questions and the vote

Answer three questions for every reply, from your own check of the images.

**Q1 EDIT FOUND.** Does the reply identify the real edit in the view it chose: both the edited object (7.1) and the
kind of change (7.2), exactly or by an allowed variant? `yes` / `no` / `unclear`.

**Q2 FABRICATED CHANGE.** Does the reply assert as fact at least one scene change that does not exist? By P2, every
asserted change other than the real edit does not exist. This includes an object said to have moved, turned, gone
missing, appeared, been replaced, recoloured or resized when it was not; a viewpoint effect presented as a change; a
rendering or physics defect; and a different object or a different instance said to have changed ("the black cup
moved" when the white cup moved). `yes` / `no` / `unclear`.

**Q3 OTHER FALSE STATEMENT.** Apart from Q2, does the reply assert as fact anything the images clearly contradict?
This includes a factual error in describing the real edit that goes beyond tolerated wording (section 8); an object
said to be visible, absent or occluded in a view where the image shows otherwise; an edited object named as an
unchanged reference (for example the other object of a swap used as a fixed landmark); a wrong count; and a false
landmark or layout fact. `yes` / `no` / `unclear`.

The vote follows mechanically, and the checker rejects any other combination:

| Q1 | Q2 | Q3 | vote |
|---|---|---|---|
| yes | no | no | TS |
| no | yes | yes or no | HALLUCINATION |
| every other combination, including any `unclear` | | | DISPUTE |

A DISPUTE vote needs at least one reason code:

- `EDIT_PLUS_FABRICATION`: Q1 yes and Q2 yes.
- `EDIT_PLUS_FALSE_STATEMENT`: Q1 yes and Q3 yes.
- `NO_EDIT_NO_FABRICATION`: Q1 no and Q2 no (vague, eliminative or unexplained).
- `TOLERANCE_BOUNDARY`: the wording sits at the edge of sections 7 and 8 and you cannot tell whether it is allowed.
- `UNCLEAR_IMAGE_EVIDENCE`: you cannot settle a claim from the images.
- `UNCLEAR_REPLY`: you cannot tell what the reply asserts.

## 7. Q1 in detail

### 7.1 The object

- **Instance:** the reply names the edited object. Names of very high semantic similarity count (mug = cup).
- **Category:** the reply names the right category without saying which instance: "the cup on the table moved" when
  the white cup moved and the black cup did not. This counts, provided the reply does not point to a different
  instance.
- **Wrong instance or wrong object:** the reply says that a specific other object or instance changed ("the black cup
  moved"). Q1 = no, and Q2 = yes.
- **Not named:** "something", "an object", "this area", with no category. Q1 = no.
- The reply points at the right place with a name that is neither the category nor a close synonym: Q1 = unclear
  (`TOLERANCE_BOUNDARY`).

### 7.2 The change: the exact kind or an allowed variant

A variant counts only when the object itself is clearly pointed out. Cite the rule ids in your record.

**move**

- M1: moved, shifted, relocated, in a different place, its position changed, now at <the correct place>.
- M2: the distance between the moved object and a nearby unchanged object increased or decreased, in the direction the
  images show.
- Not allowed: "X disappeared / was deleted / is missing" as the whole description. Q1 = unclear
  (`TOLERANCE_BOUNDARY`); if X is visible elsewhere in the same view, the absence claim is false (Q3 = yes).

**rotate**

- R1: rotated, turned, faces another way, its orientation or angle changed.
- R2: the generic "moved", "was moved", "its position changed", without an explicit claim of translation.
- Not allowed: an explicit translation that did not happen ("moved to the other end of the table", "shifted a metre
  to the left"). If the reply also notices the turning, Q1 = yes and Q3 = yes; if the translation is its only
  description, Q1 = no and Q2 = yes. Exception: a translation claim that matches `edit.assisting_translation` is true.

**swap (X and Y)**

- S1: X and Y swapped or exchanged places.
- S2: only one of them, described correctly: "X moved", "X is now where Y was", "X is in a different position".
- S3: local disappearance and appearance: "X disappeared from its spot and Y appeared there", "X was removed and Y was
  added".
- Not allowed: X said to be absent from the whole view or the whole room while the image shows X elsewhere in that
  view (Q3 = yes); the other swapped object named as an unchanged reference (Q3 = yes).

**delete (X)**

- D1: missing, gone, absent, disappeared, no longer there, deleted, removed, taken away, moved away out of the scene.
- Not allowed: a plain "X moved" (a relocation) without saying that X left. Q1 = unclear (`TOLERANCE_BOUNDARY`).
  "X moved to <place>" where no X stands: Q2 = yes.

**detail (X; see `edit.detail_kind`)**

- T1: X changed, looks different, is a different object or model, has a different colour, texture or surface.
- T2: X was replaced by another object; the old X is gone and a new one stands in its place.
- T3 (substitution only): "X disappeared / was deleted / was removed" on its own; "a new object appeared / was added"
  on its own.
- Not allowed (appearance only): X said to be gone, removed or deleted in the sense that no object stands there,
  while the recoloured or retextured X is visible there (Q3 = yes).

**mirror**

- Mi1: the room or the scene is mirrored, flipped left-right, reflected, a mirror image.
- Mi2: any existing object, or any part of the room, is mirrored or flipped left-right.
- Not allowed: anything without the mirror concept, such as individual objects relocated or turned, "seen from the
  other side", "rotated 180 degrees" or "a different room". Q1 = no. Such statements are Q2 = yes when they assert
  something the mirrored room does not show; otherwise the vote is DISPUTE (`TOLERANCE_BOUNDARY`).

**All types**

- Semantic near-equivalents count ("disappear" = delete).
- Magnitude, direction and angle are not required. If the reply states them and the images clearly contradict them,
  Q3 = yes.



## 8. Tolerated wording versus factual error

Wording tolerance never covers a factual error. The reply need not restate the generator's edit category or the whole
edit process, but it must describe the real visible change and give evidence sufficient for the view it chose.

Tolerated (not false):

- approximate positions, distances and directions;
- a loose but recognisable name for a real object (a robot called an "appliance", a square gold plate called a
  "tray");
- an object's place said to be outside a view when only a few pixels of it show at the frame edge;
- a viewpoint effect correctly described as a viewpoint effect;
- general statements about shared landmarks that the images do not contradict.

False: anything the images clearly contradict.

Asserted versus tentative: a statement that the reply raises as a possibility, checks, and then rejects or leaves open
is not an asserted fact. A statement that the reply uses to support its choice is asserted, however it is hedged.

## 9. Procedure for each packet

1. Read `facts.json`. Open all four images yourself before you read any reply. Find the edited object or objects in
   every view using the pixel hints, and record what each view shows (`edit_locate`).
2. For each reply, read the whole `visible_response`, extract its claims (section 10), and check every concrete
   visual statement against the images, including the landmarks and references the reply relies on.
3. Answer Q1, Q2 and Q3 from your checked claims, and derive the vote from the table in section 6.
4. Write the records and run the checker. Fix format problems only; never change a judgement to satisfy the checker.

## 10. Output

Write one JSON object per reply, one per line, to `review_<ROUND>.jsonl` in your packet folder, and nothing else:

```json
{
  "unit_id": "<from replies.json>",
  "packet_id": "<folder name>",
  "round": "<ROUND>",
  "seat": "<SEAT>",
  "reviewer_model": "<the exact model id stated in your own system information>",
  "choice_correct": true,
  "edit_locate": {
    "A": {"status": "visible | partly_visible | occluded | out_of_frame", "where": "<place, approx. (x, y)>"},
    "B": {"status": "...", "where": "..."},
    "C": {"status": "...", "where": "..."},
    "D": {"status": "...", "where": "..."}
  },
  "claims": [
    {
      "cid": 1,
      "quote": "<exact excerpt of visible_response>",
      "kind": "asserted_fact | tentative_hypothesis",
      "topic": "edit | other_change | unchanged_reference | visibility | landmark | count | viewpoint | render_physics | other",
      "status": "true | tolerated_wording | false | unverifiable",
      "rule_ids": ["M1"],
      "views": ["B"],
      "location": "<view, place, approx. (x, y)>",
      "evidence": "<what the images show there>"
    }
  ],
  "q1_edit_found": "yes | no | unclear",
  "q1_object": "instance | category | wrong_instance | wrong_object | not_named | unclear",
  "q1_change": "exact | allowed_variant | not_allowed | not_stated | unclear",
  "q2_fabricated_change": "yes | no | unclear",
  "q3_other_false_statement": "yes | no | unclear",
  "vote": "TS | HALLUCINATION | DISPUTE",
  "dispute_reasons": [],
  "counter_check": "",
  "item_observation": "",
  "analysis": "<3 to 6 sentences>"
}
```

- `quote` is an exact, contiguous excerpt of `visible_response`. Copy it by slicing the string in Python, never by
  retyping.
- Every claim with status `false` carries `views`, `location` and `evidence`; a human will open the image at that
  place.
- `analysis` names the edit the reply identified (or says it identified none), every false statement with the view
  and place that contradicts it, and the resulting vote.
- `item_observation` is optional and never changes Q1 to Q3 or the vote. Use it only when, after careful checking,
  you cannot find the edit where `facts.json` places it; describe what you see.
- Every record is preliminary. You never produce a final label.

===== Section 4 ROUND HEADER: R1-a through R1-d (PREPEND TO Section 3) =====

# Round R1  -  seat <SEAT> (one of four independent reviewers in this round)

You review every correct-choice reply in the packets listed below. The replies come from several solver models and
prompting conditions; both are hidden from you and are not your concern. Judge each reply on its own text. The
replies within each packet are in random order.

Claims: record every statement that decides the choice and every false statement you find. Leave `counter_check`
empty.

Packets: <list of packet folders>

===== Section 4 ROUND HEADER: S1/S2/S3, two seats per round (PREPEND TO Section 3) =====

# Strict round <S1|S2|S3>  -  seat <SEAT> (one of two independent reviewers in this round)

You have not seen these items or replies before, and you are told nothing about any earlier review of them. The
replies within each packet are in a random order that differs from any other review.

This round uses a stricter procedure, in addition to everything below:

1. Full inventory. Record every visual statement the reply asserts, not only the decisive ones, and check each one
   against the images.
2. Search pass. Search the reply for the edited object's name, its synonyms and its category, and for every change
   word (moved, shifted, turned, rotated, missing, gone, removed, appeared, swapped, replaced, changed, mirrored,
   flipped, closer, farther), so that no identification of the edit and no asserted change is missed.
3. Counter-check. Before you vote, write into `counter_check`:
   - for TS: the statement in the reply most likely to be false, and why it is not;
   - for HALLUCINATION: why no statement in the reply identifies the edit, and where in the images each fabricated
     change is contradicted;
   - for DISPUTE: exactly what a human must decide.

Packets: <list of packet folders>
\end{Verbatim}


\begin{thebibliography}{99}

\bibitem[Bai et~al.(2025)]{bai2025qwen25vl}
Shuai Bai, Keqin Chen, Xuejing Liu, Jialin Wang, Wenbin Ge, Sibo Song, Kai Dang, Peng Wang, Shijie Wang, Jun Tang, Humen Zhong, Yuanzhi Zhu, Mingkun Yang, Zhaohai Li, Jianqiang Wan, Pengfei Wang, Wei Ding, Zheren Fu, Yiheng Xu, Jiabo Ye, Xi Zhang, Tianbao Xie, Zesen Cheng, Hang Zhang, Zhibo Yang, Haiyang Xu, and Junyang Lin.
\newblock {Qwen2.5-VL} Technical Report.
\newblock \emph{arXiv preprint arXiv:2502.13923}, 2025.
\newblock \url{https://arxiv.org/abs/2502.13923}.

\bibitem[Bonnen et~al.(2024)]{bonnen2024mochi}
Tyler Bonnen, Stephanie Fu, Yutong Bai, Thomas O'Connell,
Yoni Friedman, Nancy Kanwisher, Joshua B. Tenenbaum,
and Alexei A. Efros.
\newblock Evaluating Multiview Object Consistency in Humans and Image Models.
\newblock In \emph{Advances in Neural Information Processing Systems},
volume 37, pp. 43533--43548, 2024.
\newblock \url{https://proceedings.neurips.cc/paper_files/paper/2024/hash/4cc613494d5a4f2feef21ecc229c06f5-Abstract-Datasets_and_Benchmarks_Track.html}.


\bibitem[Deitke et~al.(2020)]{deitke2020robothor}
Matt Deitke, Winson Han, Alvaro Herrasti, Aniruddha Kembhavi, Eric Kolve, Roozbeh Mottaghi, Jordi Salvador, Dustin Schwenk, Eli VanderBilt, Matthew Wallingford, Luca Weihs, Mark Yatskar, and Ali Farhadi.
\newblock {RoboTHOR}: An Open Simulation-to-Real Embodied {AI} Platform.
\newblock In \emph{Proceedings of the IEEE/CVF Conference on Computer Vision and Pattern Recognition (CVPR)}, pp. 3164--3174, 2020.
\newblock \url{https://openaccess.thecvf.com/content_CVPR_2020/html/Deitke_RoboTHOR_An_Open_Simulation-to-Real_Embodied_AI_Platform_CVPR_2020_paper.html}.

\bibitem[Deitke et~al.(2022)]{deitke2022procthor}
Matt Deitke, Eli VanderBilt, Alvaro Herrasti, Luca Weihs, Jordi Salvador, Kiana Ehsani, Winson Han, Eric Kolve, Ali Farhadi, Aniruddha Kembhavi, and Roozbeh Mottaghi.
\newblock {ProcTHOR}: Large-Scale Embodied {AI} Using Procedural Generation.
\newblock In \emph{Advances in Neural Information Processing Systems}, volume 35, pp. 5982--5994, 2022.
\newblock \url{https://proceedings.neurips.cc/paper_files/paper/2022/file/27c546ab1e4f1d7d638e6a8dfbad9a07-Paper-Conference.pdf}.

\bibitem[Dosovitskiy et~al.(2021)]{dosovitskiy2021vit}
Alexey Dosovitskiy, Lucas Beyer, Alexander Kolesnikov, Dirk Weissenborn, Xiaohua Zhai, Thomas Unterthiner, Mostafa Dehghani, Matthias Minderer, Georg Heigold, Sylvain Gelly, Jakob Uszkoreit, and Neil Houlsby.
\newblock An Image is Worth 16x16 Words: Transformers for Image Recognition at Scale.
\newblock In \emph{International Conference on Learning Representations}, 2021.
\newblock \url{https://openreview.net/forum?id=YicbFdNTTy}.

\bibitem[Fu et~al.(2021)]{fu2021front}
Huan Fu, Bowen Cai, Lin Gao, Ling-Xiao Zhang, Jiaming Wang, Cao Li, Qixun Zeng, Chengyue Sun, Rongfei Jia, Binqiang Zhao, and Hao Zhang.
\newblock {3D-FRONT}: {3D} Furnished Rooms with layOuts and semaNTics.
\newblock In \emph{Proceedings of the IEEE/CVF International Conference on Computer Vision (ICCV)}, pp. 10933--10942, 2021.
\newblock \url{https://openaccess.thecvf.com/content/ICCV2021/html/Fu_3D-FRONT_3D_Furnished_Rooms_With_layOuts_and_semaNTics_ICCV_2021_paper.html}.

\bibitem[Gan et~al.(2021)]{gan2021tdw}
Chuang Gan, Jeremy Schwartz, Seth Alter, Damian Mrowca, Martin Schrimpf, James Traer, Julian De Freitas, Jonas Kubilius, Abhishek Bhandwaldar, Nick Haber, Megumi Sano, Kuno Kim, Elias Wang, Michael Lingelbach, Aidan Curtis, Kevin Feigelis, Daniel Bear, Dan Gutfreund, David Cox, Antonio Torralba, James J DiCarlo, Josh Tenenbaum, Josh McDermott, and Dan Yamins.
\newblock {ThreeDWorld}: A Platform for Interactive Multi-Modal Physical Simulation.
\newblock In \emph{Proceedings of the Neural Information Processing Systems Track on Datasets and Benchmarks}, volume 1, 2021.
\newblock \url{https://datasets-benchmarks-proceedings.neurips.cc/paper/2021/hash/735b90b4568125ed6c3f678819b6e058-Abstract-round1.html}.

\bibitem[Hong et~al.(2026)]{hong2026esi}
Yining Hong, Jiageng Liu, Han Yin, Manling Li, Leonidas Guibas, Li Fei-Fei, Jiajun Wu, and Yejin Choi.
\newblock {ESI-Bench}: Towards Embodied Spatial Intelligence that Closes the Perception-Action Loop.
\newblock \emph{arXiv preprint arXiv:2605.18746}, 2026.
\newblock \url{https://arxiv.org/abs/2605.18746}.

\bibitem[Hu et~al.(2022)]{hu2022lora}
Edward J. Hu, Yelong Shen, Phillip Wallis, Zeyuan Allen-Zhu, Yuanzhi Li, Shean Wang, Lu Wang, and Weizhu Chen.
\newblock {LoRA}: Low-Rank Adaptation of Large Language Models.
\newblock In \emph{International Conference on Learning Representations}, 2022.
\newblock \url{https://openreview.net/forum?id=nZeVKeeFYf9}.

\bibitem[Jeong et~al.(2026)]{jeong2026xvr}
Suchae Jeong, Jaehwi Song, Haeone Lee, Hanna Kim, Jian Kim, Dongjun Lee, Dong Kyu Shin, Changyeon Kim, Dongyoon Hahm, Woogyeol Jin, Juheon Choi, and Kimin Lee.
\newblock Learning Multi-View Spatial Reasoning from Cross-View Relations.
\newblock In \emph{Proceedings of the IEEE/CVF Conference on Computer Vision and Pattern Recognition (CVPR)}, pp. 2570--2581, 2026.
\newblock \url{https://openaccess.thecvf.com/content/CVPR2026/html/Jeong_Learning_Multi-View_Spatial_Reasoning_from_Cross-View_Relations_CVPR_2026_paper.html}.

\bibitem[Khangaonkar et~al.(2026)]{khangaonkar2026inconsistencies}
Om Khangaonkar, Hadi~J. Rad, and Hamed Pirsiavash.
\newblock Multimodal Language Models Cannot Spot Spatial Inconsistencies.
\newblock \emph{arXiv preprint arXiv:2604.00799}, 2026.
\newblock \url{https://arxiv.org/abs/2604.00799}.

\bibitem[Kolve et~al.(2017)]{kolve2017thor}
Eric Kolve, Roozbeh Mottaghi, Winson Han, Eli VanderBilt, Luca Weihs, Alvaro Herrasti, Matt Deitke, Kiana Ehsani, Daniel Gordon, Yuke Zhu, Aniruddha Kembhavi, Abhinav Gupta, and Ali Farhadi.
\newblock {AI2-THOR}: An Interactive {3D} Environment for Visual {AI}.
\newblock \emph{arXiv preprint arXiv:1712.05474}, 2017.
Version 4, revised August 26, 2022.
\newblock \url{https://arxiv.org/abs/1712.05474v4}.

\bibitem[Loshchilov \& Hutter(2019)]{loshchilov2019adamw}
Ilya Loshchilov and Frank Hutter.
\newblock Decoupled Weight Decay Regularization.
\newblock In \emph{International Conference on Learning Representations}, 2019.
\newblock \url{https://openreview.net/forum?id=Bkg6RiCqY7}.

\bibitem[Ma et~al.(2025)]{ma2025threeD}
Wufei Ma, Haoyu Chen, Guofeng Zhang, Yu-Cheng Chou, Jieneng Chen, Celso de Melo, and Alan Yuille.
\newblock {3DSRBench}: A Comprehensive {3D} Spatial Reasoning Benchmark.
\newblock In \emph{Proceedings of the IEEE/CVF International Conference on Computer Vision (ICCV)}, pp. 6924--6934, 2025.
\newblock \url{https://openaccess.thecvf.com/content/ICCV2025/html/Ma_3DSRBench_A_Comprehensive_3D_Spatial_Reasoning_Benchmark_ICCV_2025_paper.html}.

\bibitem[Oquab et~al.(2024)]{oquab2024dinov2}
Maxime Oquab, Timoth\'ee Darcet, Th\'eo Moutakanni, Huy V. Vo, Marc Szafraniec, Vasil Khalidov, Pierre Fernandez, Daniel Haziza, Francisco Massa, Alaaeldin El-Nouby, Mahmoud Assran, Nicolas Ballas, Wojciech Galuba, Russell Howes, Po-Yao Huang, Shang-Wen Li, Ishan Misra, Michael Rabbat, Vasu Sharma, Gabriel Synnaeve, Hu Xu, Herv\'e Jegou, Julien Mairal, Patrick Labatut, Armand Joulin, and Piotr Bojanowski.
\newblock {DINOv2}: Learning Robust Visual Features without Supervision.
\newblock \emph{Transactions on Machine Learning Research}, 2024.
\newblock \url{https://openreview.net/forum?id=a68SUt6zFt}.

\bibitem[Park et~al.(2019)]{park2019change}
Dong Huk Park, Trevor Darrell, and Anna Rohrbach.
\newblock Robust Change Captioning.
\newblock In \emph{Proceedings of the IEEE/CVF International Conference on Computer Vision (ICCV)}, pp. 4624--4633, 2019.
\newblock \url{https://openaccess.thecvf.com/content_ICCV_2019/html/Park_Robust_Change_Captioning_ICCV_2019_paper.html}.

\bibitem[Radford et~al.(2021)]{radford2021clip}
Alec Radford, Jong Wook Kim, Chris Hallacy, Aditya Ramesh, Gabriel Goh, Sandhini Agarwal, Girish Sastry, Amanda Askell, Pamela Mishkin, Jack Clark, Gretchen Krueger, and Ilya Sutskever.
\newblock Learning Transferable Visual Models From Natural Language Supervision.
\newblock In \emph{Proceedings of the 38th International Conference on Machine Learning}, volume 139 of \emph{Proceedings of Machine Learning Research}, pp. 8748--8763, 2021.
\newblock \url{https://proceedings.mlr.press/v139/radford21a.html}.

\bibitem[Raistrick et~al.(2024)]{raistrick2024indoors}
Alexander Raistrick, Lingjie Mei, Karhan Kayan, David Yan, Yiming Zuo, Beining Han, Hongyu Wen, Meenal Parakh, Stamatis Alexandropoulos, Lahav Lipson, Zeyu Ma, and Jia Deng.
\newblock {Infinigen Indoors}: Photorealistic Indoor Scenes using Procedural Generation.
\newblock In \emph{Proceedings of the IEEE/CVF Conference on Computer Vision and Pattern Recognition (CVPR)}, pp. 21783--21794, 2024.
\newblock \url{https://openaccess.thecvf.com/content/CVPR2024/html/Raistrick_Infinigen_Indoors_Photorealistic_Indoor_Scenes_using_Procedural_Generation_CVPR_2024_paper.html}.

\bibitem[Wang et~al.(2026)]{wang2026mindcube}
Qineng Wang, Baiqiao Yin, Pingyue Zhang, Jianshu Zhang, Kangrui Wang, Zihan Wang, Jieyu Zhang, Keshigeyan Chandrasegaran, Han Liu, Ranjay Krishna, Saining Xie, Jiajun Wu, Li Fei-Fei, and Manling Li.
\newblock Spatial Mental Modeling from Limited Views.
\newblock In \emph{International Conference on Learning Representations}, 2026.
\newblock \url{https://proceedings.iclr.cc/paper_files/paper/2026/hash/858de961874816f894285ec02e50c7fe-Abstract-Conference.html}.

\bibitem[Xia et~al.(2026)]{xia2026sage}
Hongchi Xia, Xuan Li, Zhaoshuo Li, Qianli Ma, Jiashu Xu, Ming-Yu Liu, Yin Cui, Tsung-Yi Lin, Wei-Chiu Ma, Shenlong Wang, Shuran Song, and Fangyin Wei.
\newblock {SAGE}: Scalable Agentic {3D} Scene Generation for Embodied {AI}.
\newblock In \emph{Proceedings of the IEEE/CVF Conference on Computer Vision and Pattern Recognition (CVPR)}, pp. 22358--22368, 2026.
\newblock \url{https://openaccess.thecvf.com/content/CVPR2026/html/Xia_SAGE_Scalable_Agentic_3D_Scene_Generation_for_Embodied_AI_CVPR_2026_paper.html}.

\bibitem[Yang et~al.(2025)]{yang2025vsi}
Jihan Yang, Shusheng Yang, Anjali W. Gupta, Rilyn Han, Li Fei-Fei, and Saining Xie.
\newblock Thinking in Space: How Multimodal Large Language Models See, Remember, and Recall Spaces.
\newblock In \emph{Proceedings of the IEEE/CVF Conference on Computer Vision and Pattern Recognition (CVPR)}, pp. 10632--10643, 2025.
\newblock \url{https://openaccess.thecvf.com/content/CVPR2025/html/Yang_Thinking_in_Space_How_Multimodal_Large_Language_Models_See_Remember_CVPR_2025_paper.html}.

\bibitem[Yang et~al.(2026)]{yang2026mmsi}
Sihan Yang, Runsen Xu, Yiman Xie, Sizhe Yang, Mo Li, Jingli Lin, Chenming Zhu, Xiaochen Chen, Haodong Duan, Xiangyu Yue, Dahua Lin, Tai Wang, and Jiangmiao Pang.
\newblock {MMSI-Bench}: A Benchmark for Multi-Image Spatial Intelligence.
\newblock In \emph{International Conference on Learning Representations}, 2026.
\newblock \url{https://proceedings.iclr.cc/paper_files/paper/2026/hash/fe5d4bd3e2af823701b5d8f70a3c7602-Abstract-Conference.html}.

\bibitem[Yu et~al.(2026)]{yu2026mindedit}
Leyuan Yu, Xiao Tang, Minghao Liu, Xinyuan Li, Xiaokai Bai, Sheng Zhou, Qunshu Lin, Weihao Xuan, and Naoto Yokoya.
\newblock {MindEdit-Bench}: Benchmarking Object-Level Counterfactual Spatial Reasoning in {VLMs} from In-the-Wild Photos.
\newblock \emph{arXiv preprint arXiv:2607.00491}, 2026.
\newblock \url{https://arxiv.org/abs/2607.00491}.

\bibitem[Zhang et~al.(2026)]{zhang2026spinbench}
Yuyou Zhang, Radu Corcodel, Chiori Hori, Anoop Cherian, and Ding Zhao.
\newblock {SpinBench}: Perspective and Rotation as a Lens on Spatial Reasoning in {VLMs}.
\newblock In \emph{International Conference on Learning Representations}, 2026.
\newblock \url{https://proceedings.iclr.cc/paper_files/paper/2026/hash/724be4472168f31ba1c9ac630f15dec8-Abstract-Conference.html}.

\bibitem[Zhang et~al.(2025)]{zhang2025dsi}
Ziang Zhang, Zehan Wang, Guanghao Zhang, Weilong Dai, Yan Xia, Ziang Yan, Minjie Hong, and Zhou Zhao.
\newblock {DSI-Bench}: A Benchmark for Dynamic Spatial Intelligence.
\newblock \emph{arXiv preprint arXiv:2510.18873}, 2025.
\newblock \url{https://arxiv.org/abs/2510.18873}.

\end{thebibliography}
\end{document}